\documentclass[a4paper,11pt]{article}
 \usepackage[utf8]{inputenc}
 \usepackage[T1]{fontenc}
 \usepackage[margin=1in]{geometry}
 
 \usepackage{url}            %
\usepackage{booktabs}       %
\usepackage{nicefrac}       %
\usepackage{microtype}      %

\usepackage{amsmath,amsthm}
\usepackage{mathtools,thmtools}
\usepackage{mytimes}

\usepackage[colorlinks=true,allcolors=blue,hyperfootnotes=false]{hyperref}
\usepackage[capitalise]{cleveref}
\usepackage{enumitem}
\usepackage{algorithm}
\usepackage[noend]{algpseudocode}

\usepackage{xspace}
\usepackage{xcolor}

\usepackage{graphicx}
\usepackage{caption}
\usepackage{subcaption}
\usepackage{multirow}
\usepackage{array}

\newcommand{\head}[2]
  {\multicolumn{1}{>{\centering\arraybackslash}p{#1}}{\textsc{#2}}}

\usepackage{natbib}
\let\citealp\citep
\usepackage{crossreftools}
\newcommand{\ignore}[1]{}

\usepackage{marginnote}

\newcounter{TK}

\newcounter{RA}

\declaretheoremstyle[
	    spaceabove=\topsep, 
	    spacebelow=\topsep, 
        bodyfont=\normalfont,
	    headfont=\normalfont\mdseries\scshape,
    ]{theorem}

\declaretheorem[style=theorem,name=Theorem]{theorem}

\declaretheoremstyle[
	    spaceabove=\topsep, 
	    spacebelow=\topsep, 
        bodyfont=\normalfont,
        headfont=\normalfont\mdseries\scshape,
    ]{definition}

\declaretheoremstyle[
    spaceabove=\topsep, 
    spacebelow=\topsep, 
    bodyfont=\normalfont,
    headfont=\normalfont\mdseries\scshape,
    qed=$\square$, 
    ]{proofstyle}
\declaretheorem[style=proofstyle,numbered=no,name=Proof]{proof}

\declaretheorem[style=theorem,sibling=theorem,name=Lemma]{lemma}

\declaretheorem[style=theorem,numbered=no,name=Theorem]{theorem*}
\declaretheorem[style=theorem,numbered=no,name=Lemma]{lemma*}
\declaretheorem[style=theorem,numbered=no,name=Corollary]{corollary*}
\declaretheorem[style=theorem,numbered=no,name=Proposition]{proposition*}
\declaretheorem[style=theorem,numbered=no,name=Claim]{claim*}
\declaretheorem[style=theorem,numbered=no,name=Fact]{fact*}
\declaretheorem[style=theorem,numbered=no,name=Observation]{observation*}
\declaretheorem[style=theorem,numbered=no,name=Conjecture]{conjecture*}

\declaretheorem[style=definition,numbered=no,name=Definition]{definition*}
\declaretheorem[style=definition,numbered=no,name=Remark]{remark*}
\declaretheorem[style=definition,numbered=no,name=Example]{example*}
\declaretheorem[style=definition,numbered=no,name=Question]{question*}

\DeclareMathAlphabet{\mathbfsf}{\encodingdefault}{\sfdefault}{bx}{n}

\DeclareMathOperator*{\trace}{Tr}
\DeclareMathOperator*{\diag}{diag}

\newcommand{\norm}[1]{\|#1\|}

\newcommand{\Lrnorm}[1]{\mathopen{}\big\|#1\big\|}

\newcommand{\wt}[1]{\smash{\widetilde{#1}}}

\newcommand{\E}{\mathbb{E}}

\newcommand{\tr}{^{\mkern-1.5mu\scriptstyle\mathsf{T}}}

\newcommand{\st}{\star}

\newcommand{\reals}{\mathbb{R}}
\newcommand{\eps}{\epsilon}

\let\oldtfrac\tfrac
\renewcommand{\tfrac}[2]{\smash{\oldtfrac{#1}{#2}}}

\newcommand{\entofr}{en$\to$fr\xspace}
\renewcommand{\vec}{\operatorname{vec}}
\newcommand{\sfrac}[2]{\smash{#1/#2}}

 \title{Scalable Second Order Optimization for Deep Learning}

 \author{
 Rohan Anil \\
 Google Research \\
 \small\texttt{rohananil@google\!.\!com}
 \and
Vineet Gupta\\
 Google Inc \\
 \small\texttt{vineet@google\!.\!com}
 \and
 Tomer Koren\\
 Tel Aviv University and Google Research\\
 \small\texttt{tkoren@google\!.\!com}
 \and
Kevin Regan\\
 Google Inc \\
 \small\texttt{kevinregan@google\!.\!com}
  \and
 Yoram Singer\\
 Princeton University \\
 \small\texttt{y.s@cs.princeton\!.\!edu}
 }

\begin{document}
 \maketitle

\begin{abstract}
Optimization in machine learning, both theoretical and applied, is presently
dominated by first-order gradient methods such as stochastic gradient descent.
Second-order optimization methods, that involve second derivatives and/or
second order statistics of the data, are far less prevalent despite strong
theoretical properties, due to their prohibitive computation, memory and
communication costs. 
In an attempt to bridge this gap between theoretical and practical optimization,
we present a scalable implementation of a second-order preconditioned method
(concretely, a variant of full-matrix Adagrad), that along with several critical
algorithmic and numerical improvements, provides significant convergence and
wall-clock time improvements compared to conventional first-order methods on
state-of-the-art deep models. Our novel design effectively utilizes the
prevalent heterogeneous hardware architecture for training deep models,
consisting of a multicore CPU coupled with multiple accelerator units. We
demonstrate superior performance compared to state-of-the-art on very large
learning tasks such as machine translation with Transformers, language modeling
with BERT, click-through rate prediction on Criteo, and image classification on
ImageNet with ResNet-50.
\end{abstract}

\section{Introduction}

Second order methods are among the most powerful algorithms in mathematical
optimization. Algorithms in this family often use a preconditioning matrix to
transform the gradient before applying each step. Classically, the
preconditioner is the matrix of second-order derivatives (the Hessian) in
the context of exact deterministic
optimization~(e.g.,~\citealp{fletcher2013practical,lewis2013nonsmooth,nocedal1980updating}).
While second-order methods often have significantly better convergence
properties than first-order methods, the size of typical problems prohibits
their use in practice, as they require quadratic storage and cubic computation
time for each gradient update. Approximate algorithms such as quasi-Newton
methods are aimed at significantly reducing these requirements; nonetheless,
they still impose non-trivial memory costs equivalent to storing several copies
of the model (and often quadratic computation, as in the popular two-loop
recursion~\citep{nocedal1980updating}), which severely limits their use at the
immense scale of present-day deep learning.

Arguably, one of the greatest challenges of modern optimization is to bridge
this gap between theoretical and practical optimization by making
second-order methods feasible to implement and deploy at immense scale. Besides
the compelling scientific and mathematical developments it may stimulate, this
challenge has also a clear real-world significance: recent practice of training
large models suggests that the utility of common first-order methods is
quickly reaching a plateau, in large part because their time-per-step is already
negligible (compared to other parts of the computation) and cannot be optimized
further; thus, the only way to train faster is by
drastically reducing the number of steps. To this end,
second-order methods seem a very natural and promising approach.

In this paper we focus on second-order \emph{adaptive} methods for
stochastic optimization. These methods can be thought of as full-matrix
analogues of common adaptive algorithms such as
AdaGrad~\citep{duchi2011adaptive,mcmahan2010adaptive} and Adam~\citep{kingma2014adam}: they
precondition each gradient with a second moment matrix, akin to a covariance
matrix, that accumulates the outer products of the stochastic gradients.
Full-matrix versions are potentially more powerful than first-order methods as
they can exploit statistical correlations between (gradients of) different
parameters; geometrically, they can scale and rotate gradients whereas first
order methods only scale gradients. However they suffer from similar prohibitive
runtime and memory costs as Hessian-based methods.

Recent second-order methods such as the K-FAC~\citep{heskes2000natural, kfac}, K-BFGS~\citep{goldfarb2020practical} and
Shampoo~\citep{shampoo-icml} exploit the structure of deep
networks (and more generally, models described by a collection of tensors) to
mitigate the space and runtime costs of full-matrix second-order algorithms.
These methods approximate each preconditioning matrix using a factored
representation that stems from the network structure. However, in very large
applications, these algorithms are still impractical due to a number of numerical and infrastructural pitfalls, and are difficult to parallelize.

\subsection{Contributions}

We provide solutions to practical concerns and challenges that arise in
implementing and using second-order methods at large scale. Our focus will be on
the Shampoo algorithm, but most of the challenges we address are relevant to
the implementation of many other second-order methods.
\begin{itemize}%
\item 
We design and implement a pipelined version of the optimization algorithm,
critically exploiting the heterogeneity and computing power of CPU-Accelerator
coupled architectures;
\item 
We extend Shampoo in a number of ways so as to make it applicable to a larger
range of deep architectures; in particular, the extensions allow Shampoo to be
used for training very large layers such as embedding layers ubiquitous in
language and translation models;
\item 
We replace expensive spectral decompositions (e.g. SVD) used for manipulating
preconditioners with an efficient and numerically-stable iterative method for
computing roots of PSD matrices;
\item 
We describe practical challenges and limitations we faced in our design, which
we argue could be useful for the design considerations of next-generation
accelerator hardware architectures.
\end{itemize}

Our distributed implementation demonstrates
significant improvements in performance, both in terms of number of steps, and
often in actual wall-clock time, on some extremely large deep learning
tasks: %
\begin{itemize}%
\item \emph{Machine translation}: We trained Transformer models
\citep{vaswani2017attention} on the WMT'14 English to French translation task
\citep{bojar2014wmt14} in \emph{half as many steps} compared to the state-of-the-art
(well tuned Adam).
Our overall training wall-time reductions were: Transformer: 45\%
reduction ($\sim$12hrs to 6.7hrs), Transformer-Big: 37\% reduction ($\sim$47hrs
to 29.5hrs).

\item \emph{Language modeling}: We trained BERT~\citep{devlin2018bert} in
\emph{16\% fewer steps} and achieved \emph{higher masked-LM accuracy} compared to the
state-of-the-art optimizer \citep{you2019large} at 32K batch size; the \emph{overall
wall-time decreased by 4\% from 3.8 to 3.65 hours}. For this task, our system
has not yet been tuned for performance; we discuss several possible
optimizations below.

\item \emph{Click-Through Rate (CTR) prediction}: We trained the DLRM
model~\citep{naumov2019deep} on the terabyte Criteo dataset~\citep{labs_2017} at
64K batch size in \emph{half as many steps} as the current state-of-the-art
optimizer, with a \emph{wall-time reduction of 37.5\%} ($\approx\!$13mins to 8.2mins). We achieved a new
state-of-the-art performance of 80.56\% AUC ($\approx\!0.3\%$ improvement) on
this task. (An improvement of 0.1\% is considered significant; see
\citealp{rong2020distributed, wang2017deep}.) 
\end{itemize}

Finally, we showcase an implementation which already performs better in both steps to convergence and as well as wall-clock time by emulating higher precision \citep{henry2019leveraging} for ResNet-50 at 32K batch size.  We achieved the \citet{mlperf-github} target accuracy of 75.9\% \citep{mattson2019mlperf} at 32K batch size on the standard ResNet-50 ImageNet benchmark in 1729 steps which is \emph{31.7\% fewer steps} than the previous state-of-the-art \citep{nado2021large} of 2512 steps, and saw an overall \emph{13\%} reduction in wall-clock time, which can be further accelerated with better hardware/software support. An implementation in JAX \citep{jax2018github} to reproduce is available here \href{https://bit.ly/3uXXtKy}{https://bit.ly/3uXXtKy}.

One of our main points in this work was to demonstrate wall-time
speedups with second-order methods implemented on a \emph{real-world distributed
setup} being used to train state-of-the-art deep models. In our view, this is
important for influencing future hardware accelerator design and runtime
software. Indeed, first-order methods have received huge investments in tuning,
implementation, platform support and tailored accelerator hardware over the last
decade; we believe there are numerous opportunities to improve the per-step time
performance of preconditioned methods as well. 
For example, our results give a concrete justification for incorporating
64-bit accumulation units in hardware for distributed training, and further
support adding larger on-chip memory, and more (see \cref{sec:concluding-remarks}).

\subsection{Related work}

Classic techniques for addressing the high storage and computation costs of
second-order methods mostly belong to the quasi-Newton or the trust-region
families of algorithms \citep{conn2000trust,nocedal2006numerical}.
Traditionally, these methods need nearly-accurate gradients in order to
construct useful quadratic approximations and implement reliable line searches,
rendering them as suitable for training with very large batch sizes, and
resulting in expensive iterations that make the overall algorithm slow compared
with stochastic first-order methods (see, e.g.,
\citealp{bollapragada2018progressive} for a recent account).  

Our focus in
this paper is on adaptive second-order methods which are directly applicable in
a stochastic setting. That said, our effort could be relevant to quasi-Newton
and trust-region methods as well: e.g., each iteration of typical trust-region
methods amounts to solving a certain generalized eigenvalue problem, which
presents numerical difficulties of similar nature to those encountered in matrix
root/inverse computations, being addressed here. 

Various approximations to the preconditioning matrix have been proposed in the
recent literature (e.g.,~\citealp{gonen2015faster,erdogdu2015convergence,
agarwal2016second,xu2016sub,pilanci2017newton}). However, so far the only
prevalent and pragmatic approximation is the diagonal approximation.
Some recent approaches for approximating a full-matrix preconditioner are
K-FAC~\citep{kfac}, K-BFGS~\citep{goldfarb2020practical}, Shampoo~\citep{shampoo-icml} and GGT~\citep{GGT}. 
K-FAC uses a factored approximation of the Fisher-information matrix as a preconditioner, and K-BFGS uses
a similar approximation of the Hessian for a layer.
While our focus in this paper is on  Shampoo, we believe that many
of the techniques presented here could also be applied to make K-FAC practical
in large scale (see~\cref{sec:KFAC}). GGT uses a clever trick to compute a low-rank approximation to
the AdaGrad preconditioner. However, GGT maintains several hundred copies of the
gradient in memory, which is too expensive even for mid-sized models.

\citet{kfac-distr} took a first important step at experimenting with
distributed K-FAC for training deep models, using a single machine with 8 GPUs to simulate a
distributed environment for training. 
In contrast, a main thrust of our work is to demonstrate wall-time speedups with
second-order methods on a real-world distributed setup used for training
state-of-the-art deep models, that call for design considerations crucially
different than in~\citet{kfac-distr}. 
More recently, \citet{osawa2019large} scaled up K-FAC for training convolutional
networks, but fell short of reaching the accuracy of first order methods,
despite making changes to data augmentation and model architecture.

 \paragraph{Paper organization.}
 In \cref{sec:background} we
 provide some background on preconditioning methods and describe the Shampoo
 algorithm. We next discuss the various challenges one faces in a practical
 implementation of a second-order methods in~\cref{sec:challenges}, and describe
 the improvements we made to Shampoo to make it work in our system.
 In~\cref{sec:design} we describe the design of our distributed implementation
 with accelerators for deep learning. Finally, in~\cref{sec:experiments} we
 describe experiments on several datasets, showing that our implementation
 significantly outperforms common first-order methods such as SGD, Adam and
 AdaGrad, and is comparable to second order methods such as K-FAC and K-BFGS.

\section{Preliminaries}
\label{sec:background}
 \paragraph{Notation.} \label{sec:Notation}
We use lowercase letters to denote scalars and vectors, and uppercase
letters to denote matrices. $\|A\|_F$ denotes the
Frobenius norm of $A$, i.e., $\|A\|_F^2 = \sum_{i,j} A_{ij}^2$.  $A \bullet B$ denotes the Hadamard or element-wise product of $A$ and
$B$ which have the same shape, so $C = A \bullet B \iff C_{ij} = A_{ij}B_{ij}$.
$D^{\odot \alpha}$ is the element-wise power, $\smash{ (D^{\odot \alpha})_{ij} =
D_{ij}^{\alpha} }$.

We use $\preceq$ to denote the Loewner order:
given square symmetric matrices $A, B$, we write $A \preceq B$ iff $B - A$ is
positive semidefinite (PSD). 

Given a symmetric PSD matrix $A$, and $\alpha \in
\reals$, $A^\alpha$ is defined as follows: let $A = UDU\tr$ be the singular
value decomposition of $A$, where $U$ is a unitary matrix and $D$ is a
diagonal matrix (with $D_{ii} \geq 0$ as $A$ is PSD), then $A^\alpha =
UD^\alpha U\tr$, where $(D^\alpha)_{ii} = D_{ii}^\alpha$. If $\alpha < 0$, this is defined for positive definite matrices only, where $D_{ii} > 0$.

We use $\vec(A)$ to denote the flattening of the $m \times
n$ matrix $A$:  if $A$ has rows $a_1, \ldots, a_m$, then $\vec(A)$ is the $mn
\times 1$ column vector $\vec(A) = (a_1, \ldots, a_m)\tr$. 
$A\otimes B$ denotes the Kronecker product of two matrices $A$ and $B$, and we will use the identities $(A\otimes B)^\alpha = A^\alpha \otimes B^\alpha$ for $\alpha \in \reals$, and $(A\otimes B)\vec(C) =\vec(ACB\tr)$.

\paragraph{Adaptive preconditioning methods.}
\label{sec:preconditioning_methods}
First order methods iteratively update the parameters solely based on gradient
information: $w_{t+1} = w_{t} - \eta_t \bar{g}_t$ where $w_t$ and $\bar{g}_t$
are (column) vectors in $\reals^d$. Here $\bar{g}_t$ denotes a linear
combination of the current and past gradients $g_1,\ldots,g_{t}$, where
different algorithms use different combinations. Preconditioned methods take the
form $w_{t+1} = w_{t} - P_t \bar{g}_t$ where $P_t$ is an $d \times d$ matrix.
Whereas in Newton-type methods this matrix is related to the Hessian matrix of
second-order derivatives, adaptive preconditioning is based on gradient-gradient
correlations.

The parameters of a deep network are structured as a set of tensors
of order two (i.e., a matrix), three, or four. For
simplicity of presentation we focus on the matrix case---however
our design, analysis, and implementation hold for tensors of arbitrary
order. We denote the space of parameters by the matrix
$W\in\reals^{m\times n}$ and an estimate of its gradient by $G$.
Full matrix Adagrad flattens $W, G$ to vectors $w, g$
of dimension $mn$, it thus requires $m^2n^2$ space to store the preconditioner and
$m^3n^3$ time to perform the update. $m$ and $n$ can be as large as $10^4$ in
large models, thus rendering full-matrix preconditioning
impractical. For this reason, both AdaGrad and Adam constrain the
preconditioning matrices to be diagonal.
Shampoo bridges the gap between full matrix preconditioning
and the diagonal version by approximating the full matrices by a Kronecker product.

\paragraph{The Shampoo algorithm.}
\label{sec:shampoo}

We describe Shampoo in the context of the Online Convex Optimization (OCO)
framework, which generalizes stochastic optimization~(see, e.g.,
\citealp{shalev2012online,hazan2016introduction}). 
In OCO, learning progresses in rounds where on round $t$ the learner receives an
input $X_t$ and then uses the parameters $W_t$ to form a prediction
denoted $\hat{y}_t$. After making the prediction, the true outcome $y_t$ is
revealed. The discrepancy between the true and predicted outcomes is assessed by
a loss function $\ell$ which takes values in $\reals_+$. The learner then uses
the discrepancy to update the matrix to $W_{t+1}$ and prepare for the next
round. For instance, the input on round $t$ can be an example $x_t\in\reals^k$
for which the learner predicts $\hat{y} = f(W_t, x_t)$ where
$f:\reals^k\rightarrow\reals$ and the loss is a function $\ell: \reals \times
\reals \rightarrow \reals_+$ such as $\ell(\hat{y},y) = (y - \hat{y})^2$ or
$\ell(\hat{y},y) = \log(1+\exp(-y\hat{y}))$.

Stochastic gradient methods use the loss gradient
$G_t = \nabla_W \ell(f(W,x_t),y_t)$, thus 
$G_t\in\reals^{m\times{}n}$ if the parameters are shaped as a matrix $W \in \reals^{m\times n}$. For matrix-shaped parameters, Shampoo
tracks two statistics over the course of its run, $L_{t}$ and
$R_{t}$, which are defined as follows:
\begin{equation*}
	L_t = \epsilon I_m + \textstyle\sum_{s=1}^t G^{}_s G_s\tr~;
	\qquad
	R_t = \epsilon I_n + \textstyle\sum_{s=1}^t G_s\tr G^{}_s 
	~.
\end{equation*}
Note that $L_t\in\reals^{m\times m}$ and $R_t\in\reals^{n\times n}$.
The full matrix Adagrad preconditioner $H_t$ can be approximated as $(L_t \otimes R_t)^{1/2}$.
Thus the Adagrad update rule $w_{t+1} = w_{t} - \eta H_t^{-1/2} g_t$  yields the 
Shampoo update rule for the parameter matrix $W$:
\begin{align*}
  W_{t+1}  &= W_t - \eta\, L_t^{-1/4} G_t R_t^{-1/4} ~.
\end{align*}

\section{Scaling-up Second Order Optimization}
\label{sec:challenges}

As described earlier, second order methods have not become the predominant method of choice for large scale deep learning as they come with several algorithmic, numeric and infrastructure challenges. We discuss these challenges and design choices in the development of the distributed implementation of Shampoo. 

These challenges arise from the fact that
modern accelerators are highly optimized for training using first-order
optimizers, which have low computational and memory requirements. The Shampoo
algorithm is computationally expensive. The extra computation in Shampoo
compared to standard first-order methods is in the following steps:
\begin{itemize}%
  \item Preconditioner statistics computation:
		$L_t = L_{t-1} + G_t G_t\tr \; \mbox{ and } \; R_t = R_{t-1} + G_t\tr G_t ~;$
  \item Inverse $p$'th root computation:
		$L_t^{-1/4} \mbox{ and } R_t^{-1/4} ~; $
  \item Preconditioned gradient computation:
		$L_t^{-1/4} G_t R_t^{-1/4} ~.$
\end{itemize}

Preconditioner statistics and gradient computations are expensive for large fully connected as well as embedding layers, we address these below. 
Computing the inverse  $p$'th roots is slow ---as much as 100 times the step time in some cases---and calculating these without slowing down training was a key challenge in our system.

\subsection{Preconditioning of large layers}

Modern ML architectures often use very large embedding layers, where the longer
dimension can be in the millions. For example, DLRM \citep{naumov2019deep} on
Criteo-1Tb uses a vocabulary with $\sim\!\!186$ million hash buckets, while in
Transformer models \citep{shazeer2018mesh} the largest layer can have up to
65536 units {\em per} dimension. This makes preconditioning impossible due to
$O(d^2)$ memory and $O(d^3)$ computational complexity.  We show how to extend
Shampoo to overcome these problems; we provide proofs and convergence results in
\cref{sec:Proofs}.

\paragraph{Large layers.}\label{subsec:large_tensors}
For embedding layers specifically, we extend the Shampoo algorithm to allow us to use only one of the preconditioners,
in case both preconditioners are too expensive to compute. Our choice is empirically supported by the
experiments shown in~\cref{fig:en_fr_latency,fig:en_fr_preconditioning,fig:criteo} which
suggest that there is a benefit from preconditioning one dimension of the large softmax and
embedding layers with minimal increase in time. The following result
allows us to choose a subset of preconditioners:%

\begin{lemma}\label{lemma:extend}
	Let $G_1, \ldots, G_t \in \reals^{m\times n}$ be matrices of rank at
	most $r$. Let $g_s = \vec(G_s)$ and define $\widehat{H}_t = \epsilon I_{mn} + \sum_{s =1}^t g_s g_s\tr ~.$
	Let $L_t, R_t$ be defined as above: $L_t =  \epsilon I_{m} + \sum_{s =1}^t G_s G_s\tr ~, R_t =
	\epsilon I_{n} + \sum_{s =1}^t G_s\tr G_s ~.$ Then for any $p, q > 0$ such that $\sfrac{1}{p} + \sfrac{1}{q} = 1$,
			we have $\widehat{H}_t \preceq rL_t^{\sfrac{1}{p}} \otimes  R_t^{\sfrac{1}{q}}$.
\end{lemma}

A consequence is that for any $p, q > 0$ such that $\sfrac{1}{p} + \sfrac{1}{q}
= 1$, the full AdaGrad preconditioned gradient
$\smash{\widehat{H}_t^{-\sfrac{1}{2}}g_t }$ is approximated by
$(L_t^{\sfrac{1}{p}} \otimes  R_t^{\sfrac{1}{q}})^{-\sfrac{1}{2}}g_t$, giving us
$\smash{ \wt{G}_t  = L_t^{-\sfrac{1}{2p}} G_t R_t^{-\sfrac{1}{2q}} }$. Now, by
choosing $(p,q)=(1,\infty)$ and $(p,q)=(\infty,1)$ we obtain the simple
preconditioned gradients: $\smash{ G^{}_t R_t^{-\sfrac{1}{2}} }$ and
$\smash{L_t^{-\sfrac{1}{2}} G^{}_t }$. \cref{thm:regret-2d} shows
that~\cref{lemma:extend} can be used to prove a regret bound for this extended
Shampoo in the online convex optimization setting---this provides intuitive
justification for the usefulness of this approximation. We further optimize the
computation of these preconditioned gradients for embedding layers by taking
advantage of the sparse inputs (details in \cref{sec:Embedding}).

\paragraph{Preconditioning blocks from large tensors.} \label{subsec:blocked_tensors}
In addition to embedding layers, large models occasionally have large fully
connected layers. To reduce the computational cost of computing statistics and
preconditioned gradient: we divide the tensor into blocks
and treat each individual block as a separate tensor. Concretely this entails
dividing a tensor $W\in\reals^{km\times kn}$, into $W_{1,1} \ldots W_{m,n}$ such
that $W_{i,j}\in\reals^{k\times k}$ $\forall i,j $. Shampoo still converges in this case
in the convex setting (\cref{thm:regret-block}), showing that the extension is justified.

\begin{lemma}\label{lemma:extendblock} 
Assume that $g_1, \ldots, g_t \in \reals^{mk}$ are vectors, and let $g_i =
[g_{i,1}^{}, \ldots, g_{i,k}^{}]$ where $g_{i,j}^{} \in \reals^m$. Define
$\widehat{H}_t =  \epsilon I_{mn} + \sum_{s =1}^t g_s^{} g_s\tr$, and let $B_t
\in \reals^{mk\times mk}$ be the block diagonal matrix with $k$ $m \times m$
blocks, where the $j$-th block is $B_t^{(j)} =  \epsilon I_{m} + \sum_{s =1}^t
g_{s,j}^{} g_{s, j}\tr ~.$ Then $\widehat{H}_t \preceq k B_t$. 
\end{lemma}

We performed experiments to study the effect of partitioning intermediate layers
into blocks, in which we observed that the latter had minimal impact on quality
of the solution while providing faster step time as well as reduced memory
overheads; see \cref{fig:en_fr_blocked}.

\paragraph{Delayed preconditioners.}
As remarked above, computing the preconditioners is the most expensive
computation in every Shampoo step. In \cref{fig:en_fr_delay} we show that we can
compute the preconditioners once every few hundred steps without a significant
effect on the accuracy which indicates that the loss function landscape does not
change significantly with each step. We observe that there is a performance/quality
tradeoff here --- in our experiments we set the frequency of computing preconditioners
to the smallest value that does not degrade performance, i.e. the number of training
steps that can be completed in the amount of time needed to compute the largest preconditioner.
The only way to increase the frequency of computing preconditioners is with better
hardware/software support.

\subsection{Roots of ill-conditioned matrices}

Inverse $p$'th roots (where typically $p = 2, 4, 8$) can be computed using SVD,
but there are efficient iterative algorithms such as the coupled Newton iteration
algorithm~\citep{guo2006schur,iannazzo2006newton} that can compute the inverse $p$'th root via a
sequence of matrix-vector and matrix-matrix products, which are highly optimized
on modern accelerators. However, our experiments suggest that on real workloads
the condition numbers of the $L_t, R_t$ matrices are very large (see
\cref{fig:condition_numbers} in \cref{sec:SNopt}) so both SVD and the coupled iteration must be run in
double-precision, but this is very expensive on accelerators. We applied several
further optimizations to speedup the coupled Newton iteration in our implementation; these are
described in \cref{sec:SNopt}.

\subsection{Deploying on current ML infrastructure}

\begin{figure*}[ht!]
\begin{center}
\vspace{-0.25cm}
\includegraphics[width=0.80\textwidth]{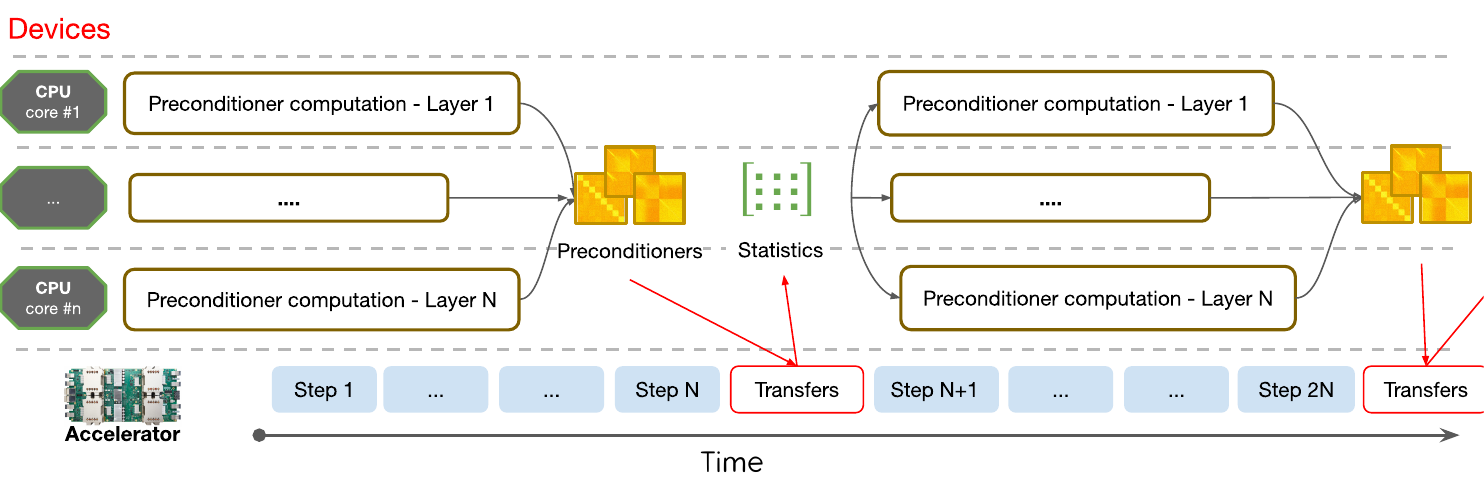}
\vspace{-0.5cm}
\end{center}
\caption{\small Timeline illustrating the design of the optimization algorithm.
	Preconditioner statistics ($L_t$ and $R_t$) are computed
	at each step by the accelerators. Preconditioners ($L_t^{\sfrac{1}{4}}$, $R_t^{\sfrac{1}{4}}$) are computed every $N$ steps and this
	computation is distributed to all available CPU cores.
	}
\label{fig:system}
\end{figure*}
\paragraph{Heterogeneous training hardware.}
 Neural network accelerators are custom designed to run machine learning workloads faster and at lower cost. Accelerator design is trending towards preferring lower-precision arithmetic that satisfy both of these goals on existing benchmarks. We find that we need double-precision arithmetic for many layers in these models as described above, which makes running computation on accelerators relatively expensive, and therefore we had to design the system to leverage the existing underutilized CPUs attached to the accelerators (\cref{sec:design}). Note that for the ResNet-50 experiments, we used single-precision arithmetic via emulation \citep{henry2019leveraging} and with sublayer blocked preconditioning with dimension 128 to significantly cut down the cost of the inverse.

\paragraph{API inflexibility.}
Deep learning libraries such as TensorFlow \citep{tensorflow} offer APIs for
optimizer implementation that are well suited for first-order optimizers and for
mini-batch training. However our design requires that we interact with the training loop
in non-standard ways as we need to pipeline the preconditioner computations --- this requires framework level changes. 

Our Transformer experiments were carried out in the Lingvo~\citep{lingvo} TensorFlow framework, while BERT-Large, DRLM, and ResNet-50 used the MLPerf v0.7 Tensorflow open source competitive baselines \citep{mattson2019mlperf}. Experimentation required changes to the training loop such as gathering statistics at regular intervals, distributing computation across all the CPUs available in the cluster without blocking the TPU training, as well as updating the preconditioners. We anticipate our work will encourage the development of more flexible API's in machine learning libraries to fully utilize heterogeneous hardware.

\section{Distributed System Design}
\label{sec:design}

We present our distributed system design of the modified Shampoo algorithm. Our
method is designed to run effectively on modern neural network accelerators such
as TPUs~\citep{jouppi2017datacenter} or GPUs. We first describe the standard
data parallelism paradigm used in training models on these
accelerators~\citep{dean2012}. Parameters are replicated on each core of the
accelerator, and each core computes forward propagation and back propagation on
a sub-batch (a subset of a mini-batch, which itself is a small randomly selected
subset of the training set) of input examples. These gradients are averaged
across all cores via all-reduction to get the average gradient for the
mini-batch. Each core uses the average mini-batch gradient to update its copy of
the parameters.

All-reduction adds a barrier as all the cores need to synchronize to compute the
mini-batch gradient. \cref{fig:en_fr_latency} shows the overhead of each of the
steps on a Transformer~\citep{vaswani2017attention} described in the experiment
section. We observe that the overheads from all-reduction and weight updates are
a minor part ($<\!5\%$) of the overall step time.

The overall design of our implementation is illustrated by the timeline in
\cref{fig:system}. As discussed in the previous section the preconditioner
computation (inverse $p$th root) is expensive and requires double precision,
also we need to do this computation once every few hundred steps. These
observations naturally suggested using the often underutilized CPUs on the
machines to which the accelerators such as GPUs or Cloud TPUs are attached. 
CPUs offer double precision arithmetic while being cheaper which makes them a perfect choice to run the preconditioner computation without
adding any extra overhead to the training, as the computation is pipelined and
runs asynchronously without blocking the training loop. 

Preconditioners need to be computed for every layer of the network so we
distribute the computation across all the CPUs that are part of the training
system. As a result, the most expensive step in Shampoo adds almost nothing to
the overall training time. Moreover, the computational overhead of
preconditioned gradient is independent of the
batch size. Thus, increasing the batch size allows us to linearly decrease the
overhead making Shampoo practical for very large scale training setups. On smaller problems (e.g., CIFAR-10, see \cref{sec:cifar-details}), we find that our design still results in training time improvements as preconditioner computations take very little time.

\section{Experiments} \label{sec:experiments} We compare our method against
various widespread optimization algorithms for training large state-of-the-art
deep models for machine translation, language modeling, recommendation systems
as well as image classification. Full details of the experiments and hyperparameter tuning are given in
\cref{sec:exper-details}.

\subsection{Comparison of second order methods}

\begin{figure}[htb]
  \centering
  \begin{tabular}{ccc}
  \includegraphics[width=0.3\textwidth]{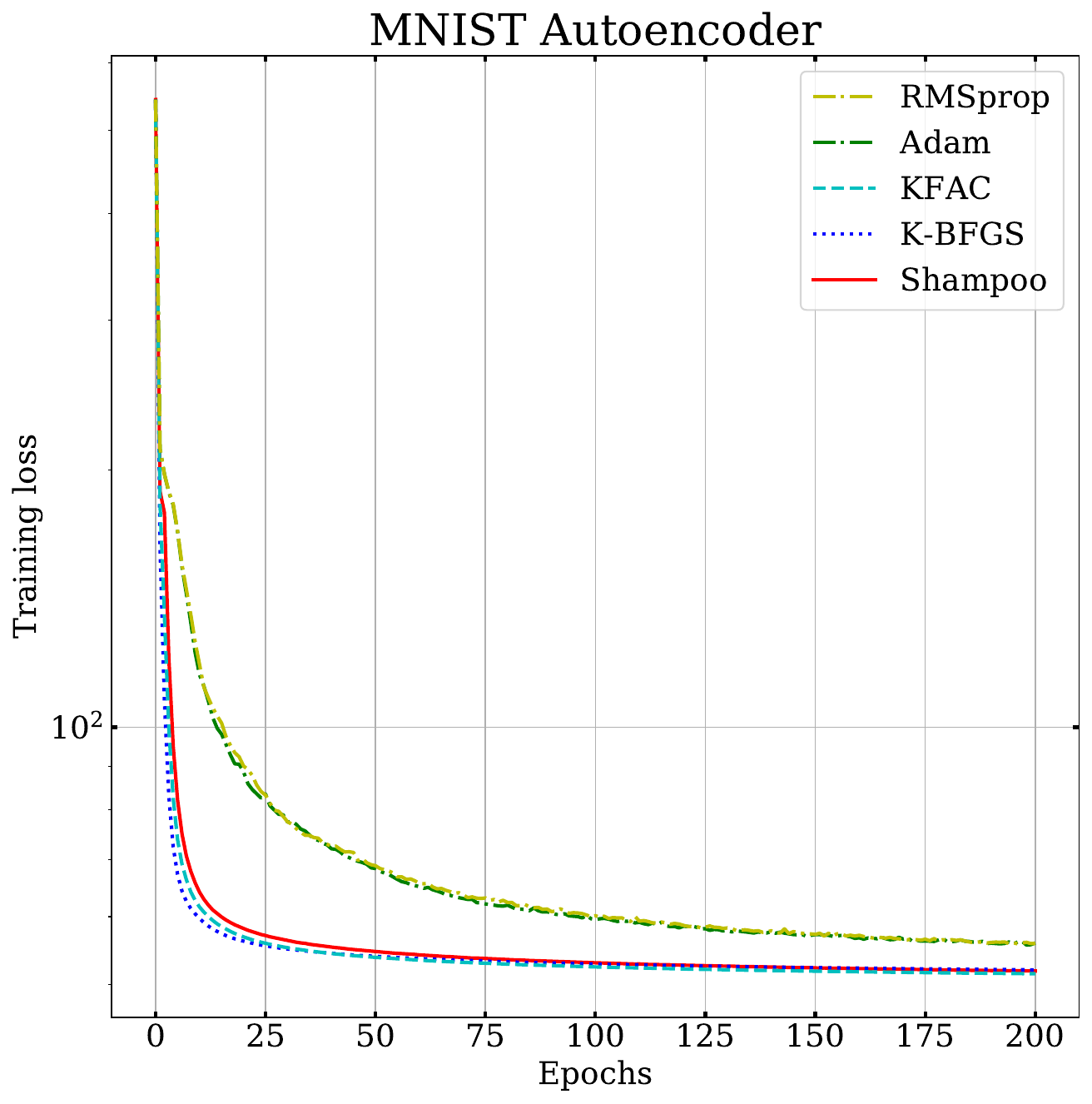}&
  \includegraphics[width=0.3\textwidth]{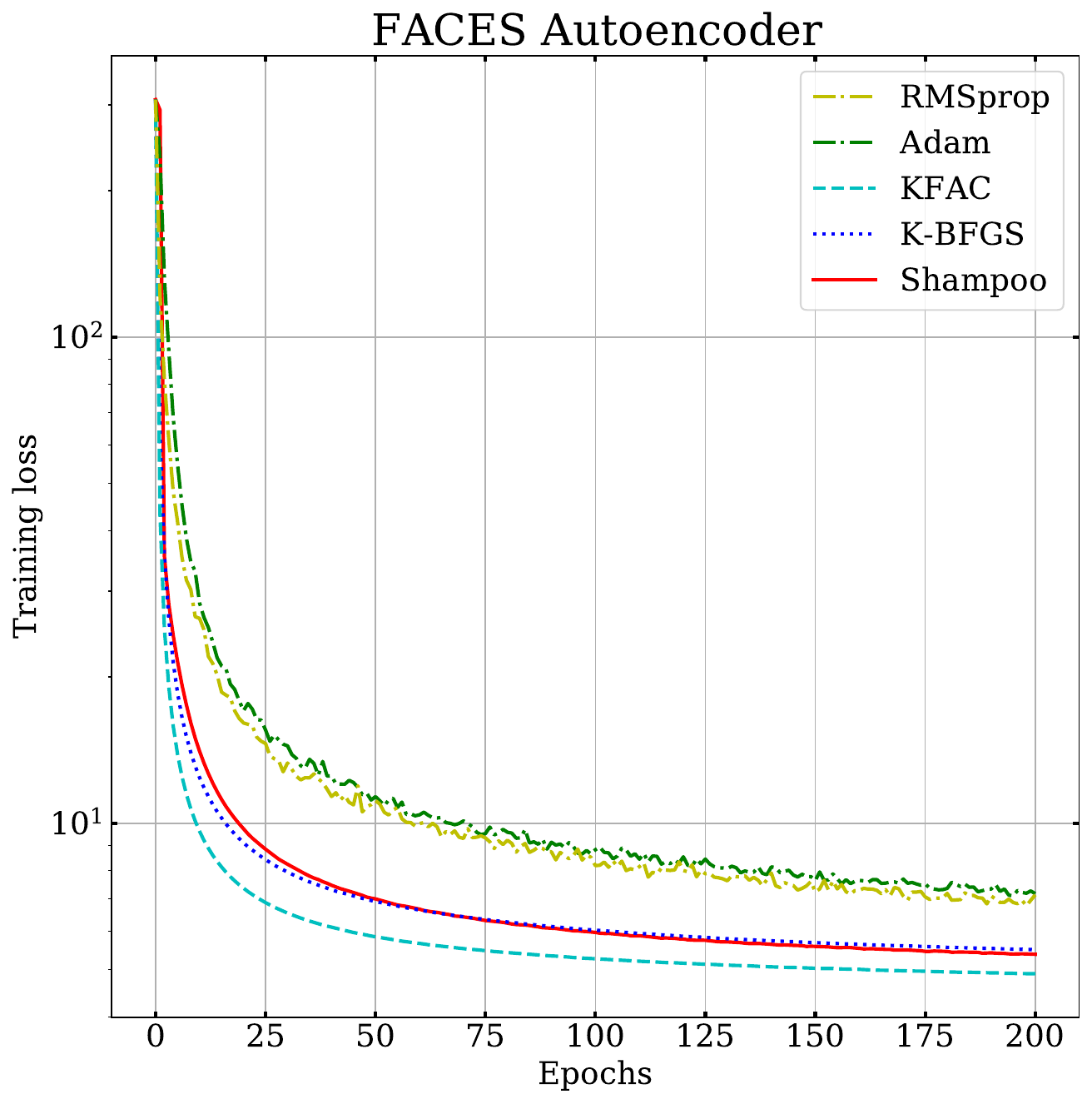}&
  \includegraphics[width=0.3\textwidth]{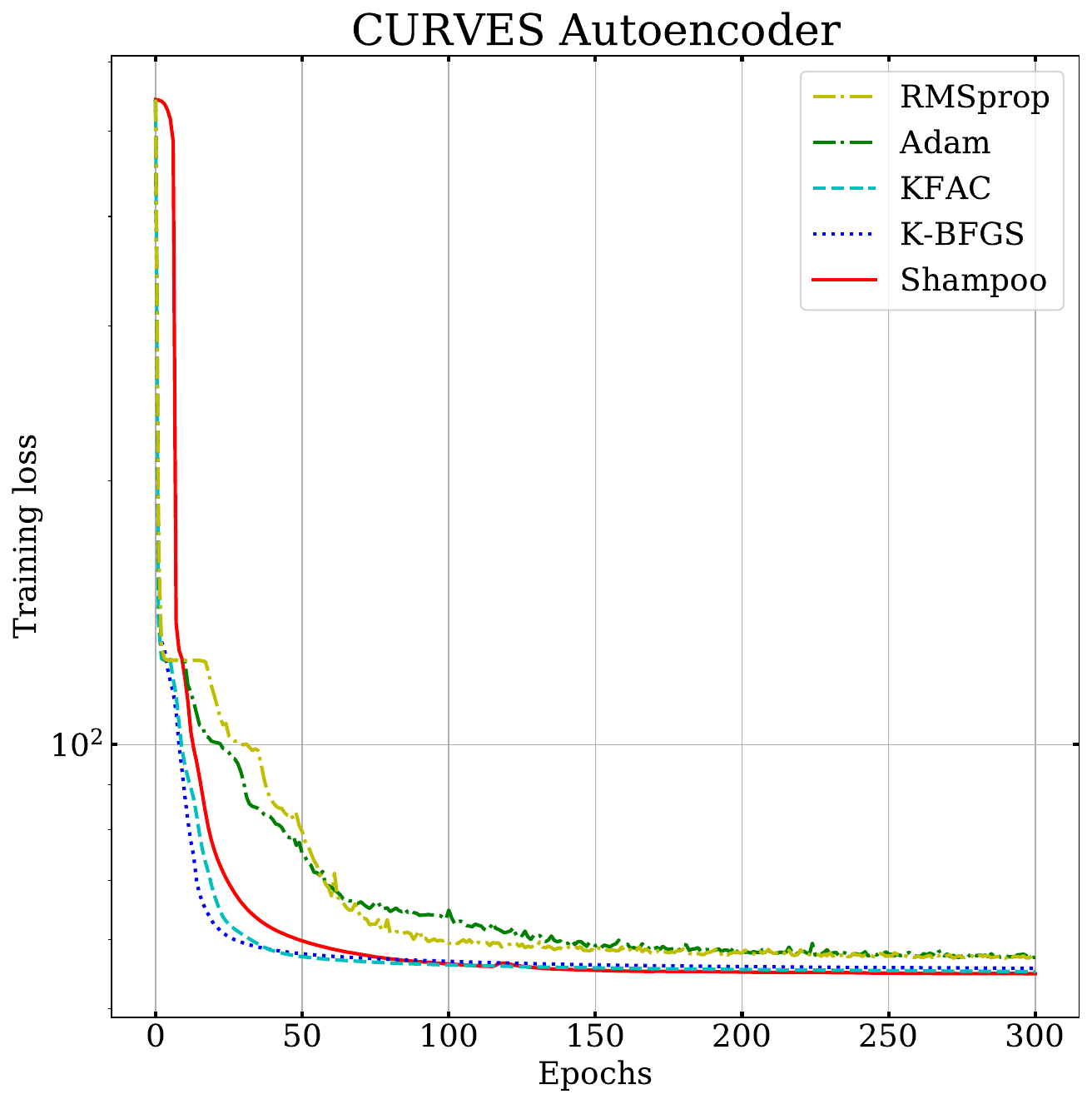}\\
  \includegraphics[width=0.3\textwidth]{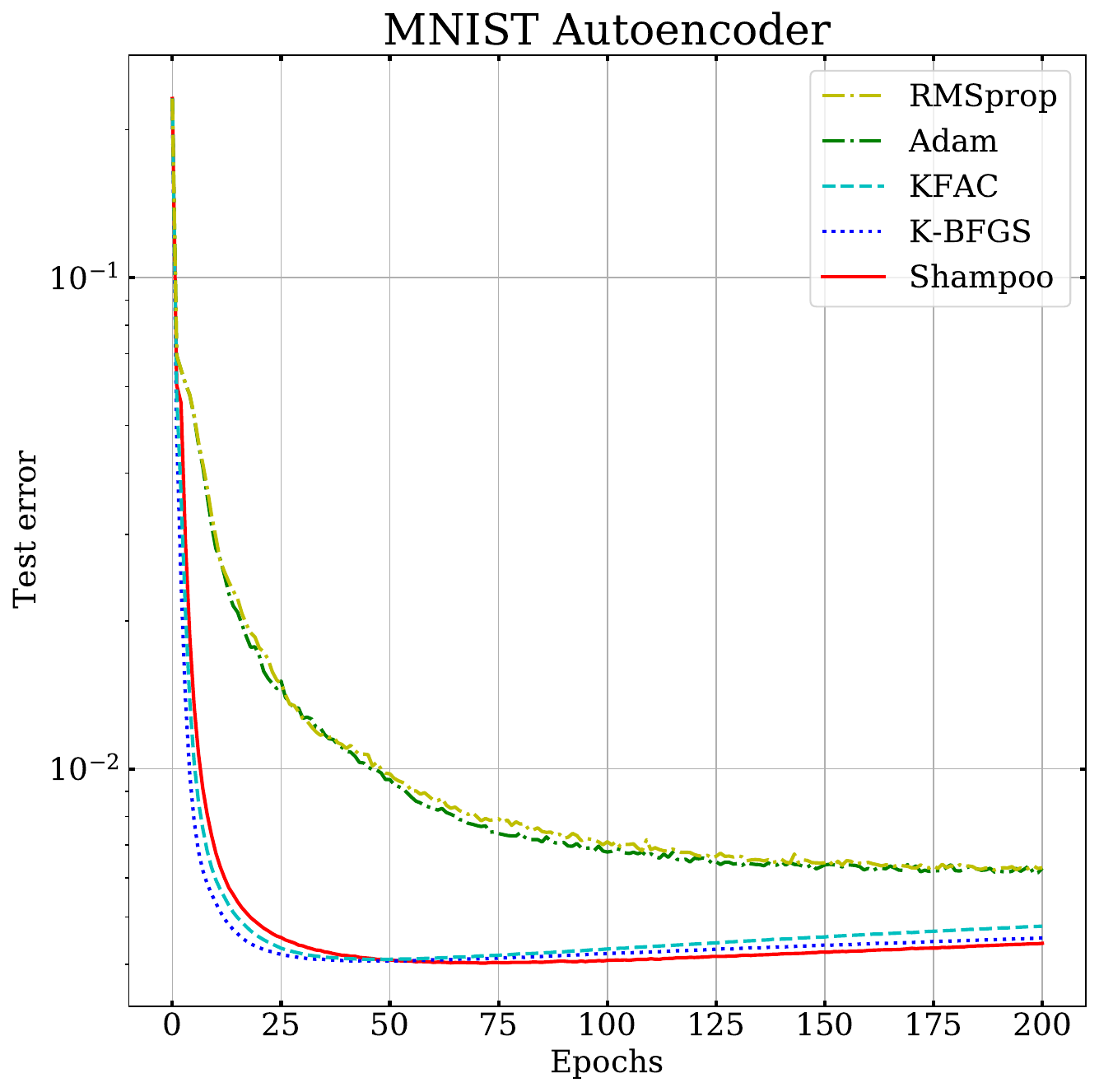}&
  \includegraphics[width=0.3\textwidth]{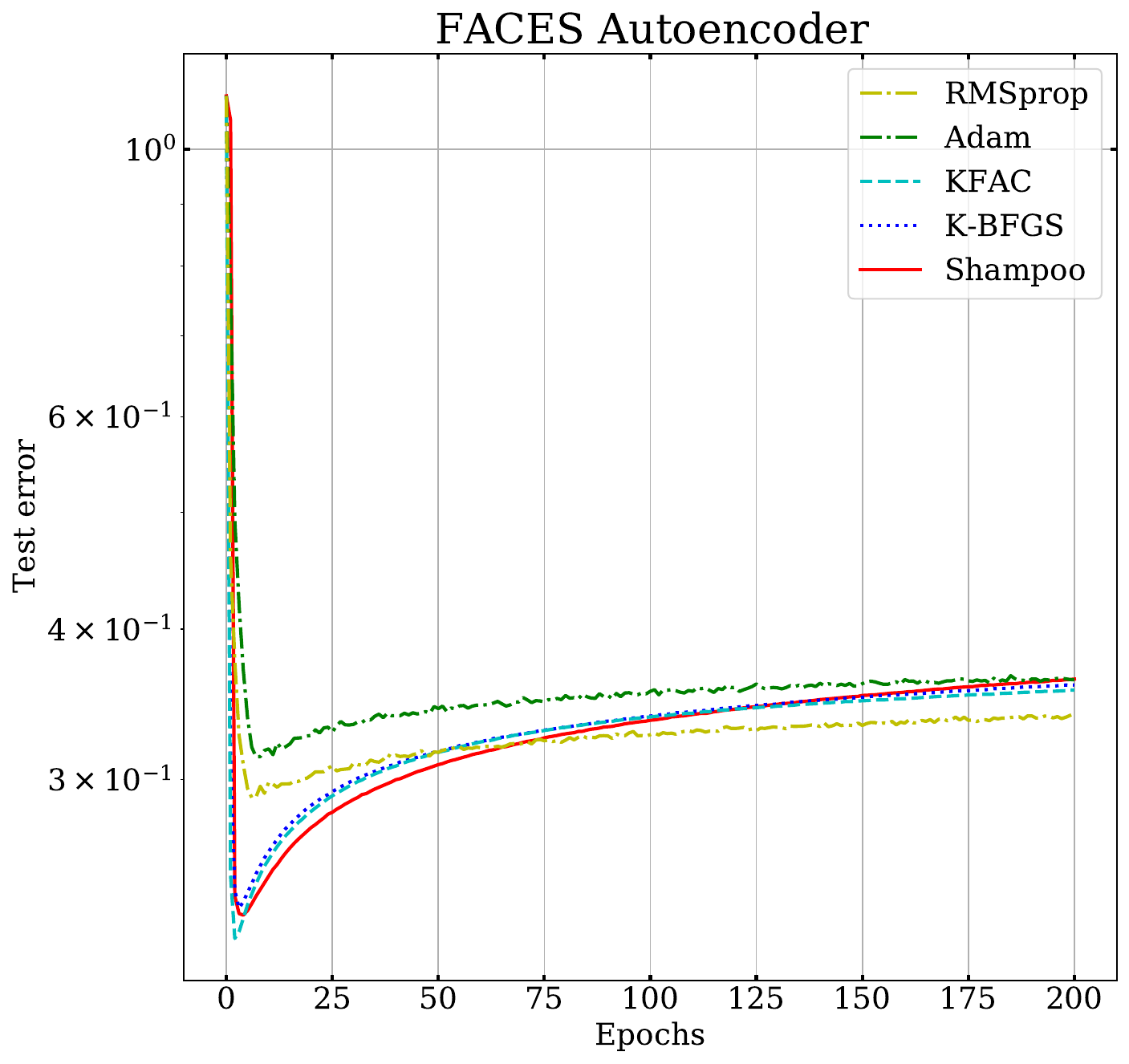}&
  \includegraphics[width=0.3\textwidth]{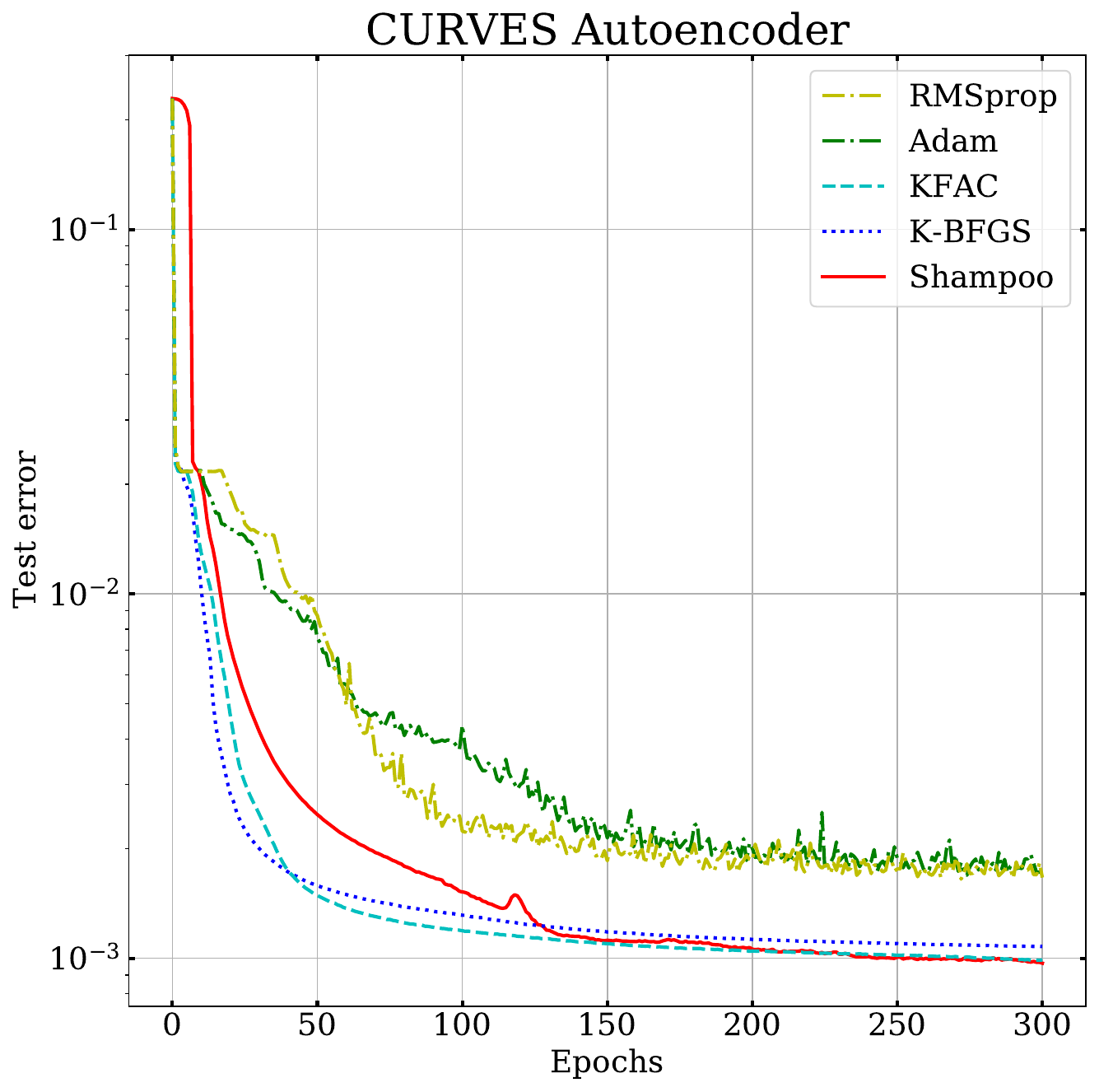}
\end{tabular}
  \caption{\small{Training losses and test errors for various optimizers for the MNIST, FACES and CURVES autoencoder tasks.}}
  \label{fig:autoencoders}
\end{figure}

We compared Shampoo with KFAC \citep{kfac} and K-BFGS \citep{goldfarb2020practical} for standard autoencoder tasks on
MNIST, FACES and CURVES, and found that all second order algorithms performed
approximately the same, and far better than first order optimizers.
\cref{fig:autoencoders} shows the training losses and test errors on these autoencoder
tasks; see~\cref{sec:second-order-experiments} for complete details on the
experiments. Scaling up each of these second order methods to work
on state-of-the-art deep networks at scale is both a research and engineering
challenge; we leave that for future work, and instead focus on comparison of
Shampoo with existing baselines based on well-tuned first order methods in a
variety of tasks. We used PyTorch \citep{NEURIPS2019_9015} code available from \citep{goldfarb2020practical} for the benchmarking.

\subsection{Machine Translation with a Transformer}

\begin{figure}[ht]
  \centering
  \begin{subfigure}{0.45\textwidth}
  \centering
  \includegraphics[width=1.0\textwidth]{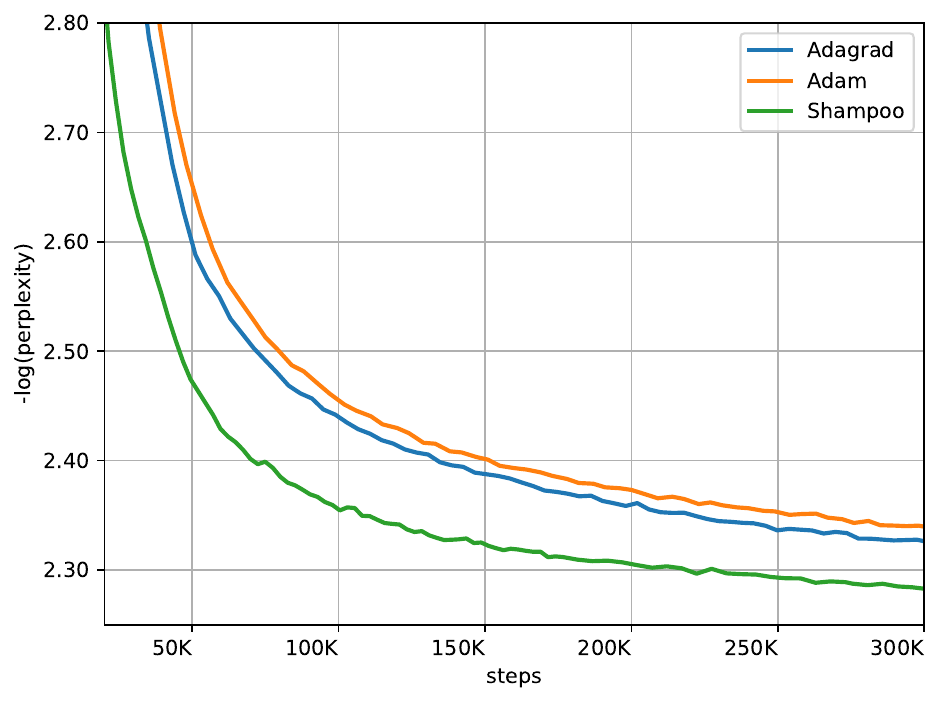}
  \vspace{-0.5cm}
  \caption{\label{fig:en_fr_perplexity}}
  \end{subfigure}
  \begin{subfigure}{0.47\textwidth}
  \centering
  \includegraphics[width=1.0\textwidth]{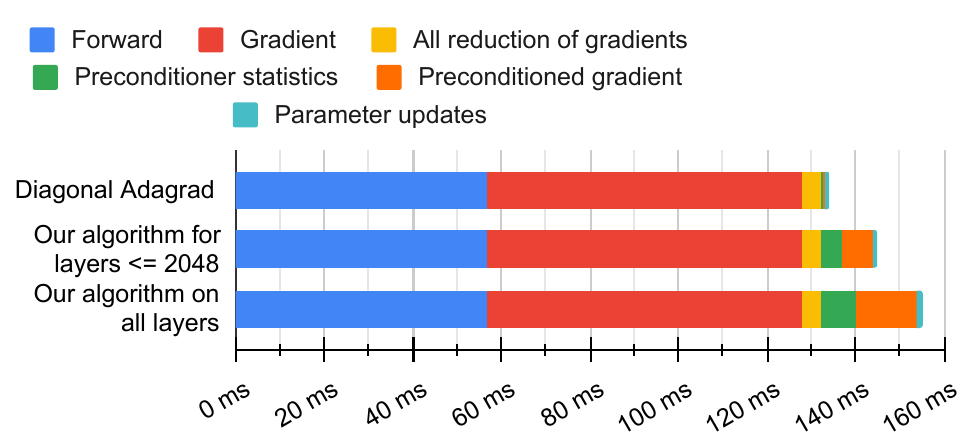}
  \caption{\label{fig:en_fr_latency}}
  \end{subfigure}
  \vspace{-0.25cm}
  \caption{\small{Results for a Transformer model on WMT'14 \entofr, trained
  with batch size of 1536. (Top) Test log-perplexity vs.~number of steps; the
  algorithm converges 1.95x faster in steps, while being only $\approx\!16\%$
  slower per step. This allows the method to attain a particular log-perplexity
  in \emph{40\% less wall-time}. (Bottom) Detailed breakdown of latency of a
  single step (\cref{sec:steptime}). Diagonal AdaGrad optimizer: 134ms, Shampoo:
  145ms (all layers except embedding and softmax layers) and 155ms (all layers).
  Preconditioner computation is pipelined and distributed over CPUs, thus does not
  add any overhead, and the transfer latency ($\approx$100ms) is amortized over
  hundreds of steps. }}
  \vspace{-0.25cm}
\end{figure}

We demonstrate the effectiveness of our implementation on the standard machine translation dataset from WMT’14 English to French (\entofr) with 36.3M sentence pairs. We used the state-of-the-art Transformer architecture \citep{vaswani2017attention}. This architecture contains 93.3M parameters and consists of 6 layers for its encoder and decoder. Each layer is composed of 512 model dimensions, 2048 hidden dimensions, and 8 attention heads. The model makes use of a sub-word vocabulary that contains 32K word pieces~\citep{schuster12}. The experiment was run on 32 cores of a Cloud TPU v3 Pod, and the implementation of the optimizer was carried out in the
Lingvo~\citep{lingvo} framework. Our results are shown in~\cref{fig:en_fr_perplexity}: our algorithm achieves the same accuracy as AdaGrad or Adam in about half as many steps.

\begin{figure*}[ht]
  \centering
  \begin{subfigure}{0.30\textwidth}
  \centering
  \includegraphics[width=1.0\linewidth]{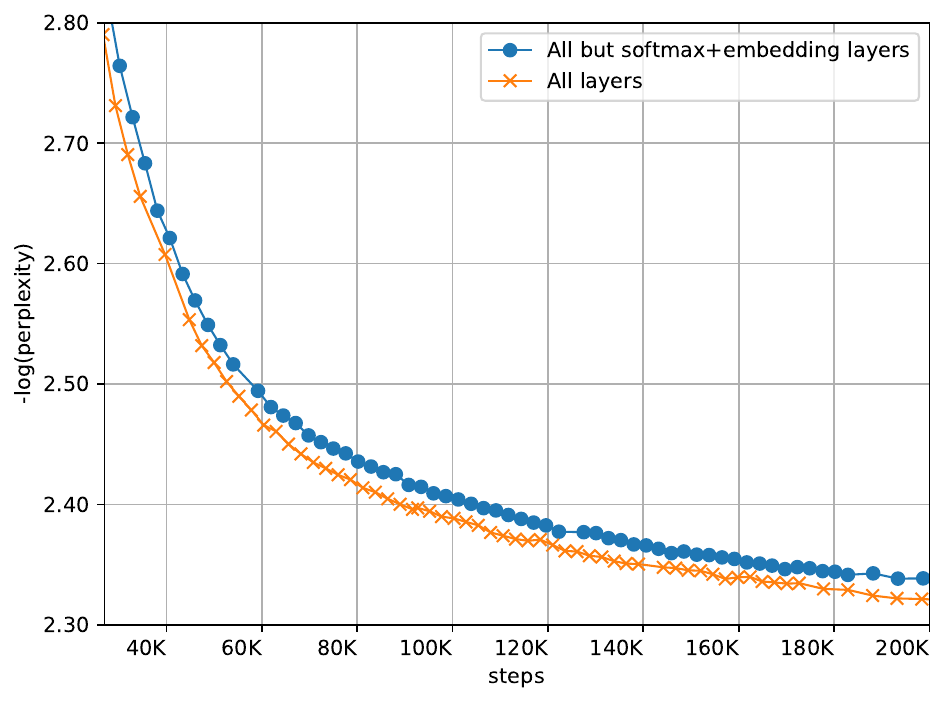}
  \caption{}
  \label{fig:en_fr_preconditioning}
  \end{subfigure}
  \begin{subfigure}{0.30\textwidth}
  \centering
  \includegraphics[width=1.0\linewidth]{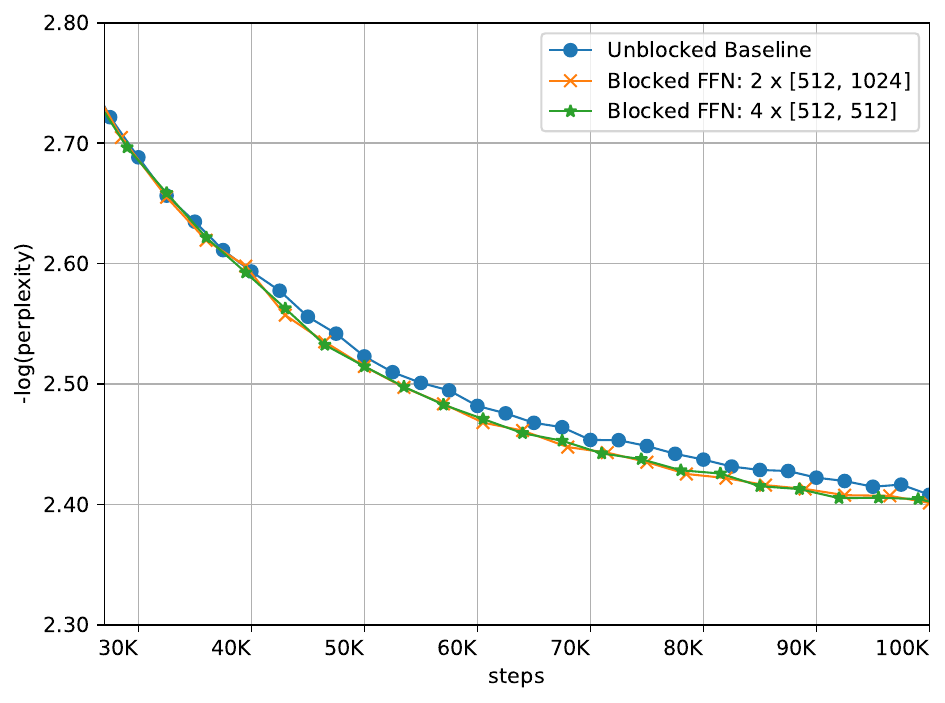}
  \caption{}
  \label{fig:en_fr_blocked}
  \end{subfigure}
  \begin{subfigure}{0.30\textwidth}
  \centering
  \includegraphics[width=1.0\linewidth]{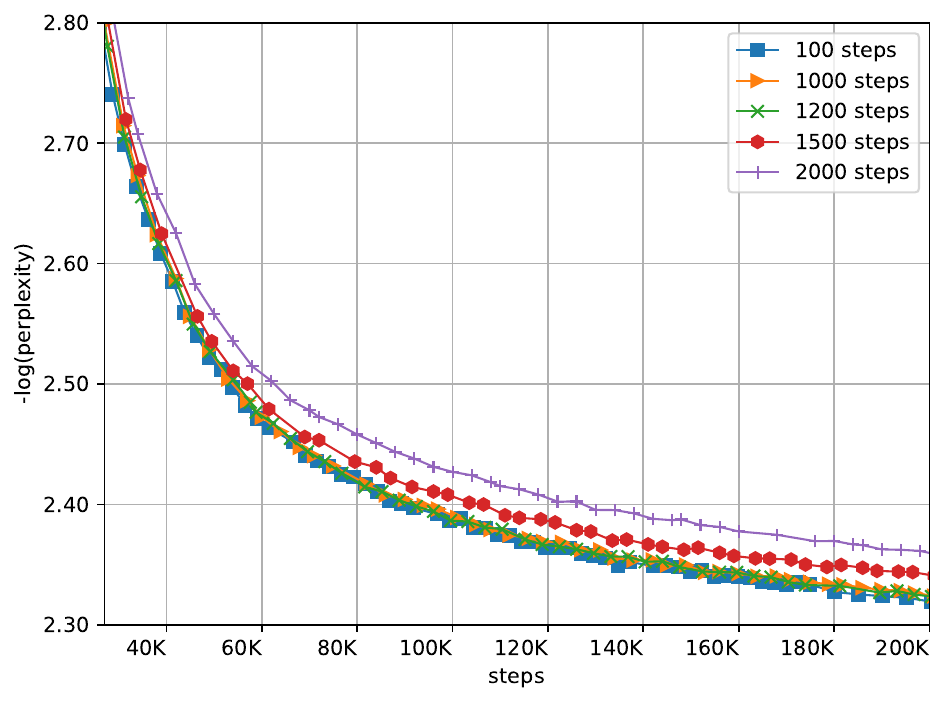}
  \caption{}
  \label{fig:en_fr_delay}
  \end{subfigure}
  \vspace{-0.25cm}
  \caption{\small{Impact of Shampoo extensions on WMT'14 \entofr training: (a)
  preconditioning applied to all layers except embedding and softmax layers,
  vs.~applied to all layers;
  (b) preconditioning with fully-connected layers partitioned into sub-blocks;
  (c) varying interval between preconditioner updates.}}
\end{figure*}

\paragraph{Preconditioning of embedding and softmax layers.}
Following the first extension in \cref{subsec:large_tensors} the
algorithm preconditions the large layers with only one of the preconditioners
($\smash{G_tR_t^{-1/2}}$ or $\smash{L _t^{-1/2}G_t}$) to make it tractable.
\cref{fig:en_fr_latency} shows the increase in step time is only 6\% while
\cref{fig:en_fr_preconditioning} shows that we can reduce the number of steps to
convergence by $\approx\!20\%$.

\paragraph{Reducing overhead in fully-connected layers.}
Following the second extension in \cref{subsec:large_tensors} we ran two experiments where we partitioned fully connected layer of size [512, 2048] into two blocks of size [512, 1024] and four blocks of size [512, 512]. Our experiments show no drop in quality with a small reduction in runtime ($<\!3\%$).

\subsection{Transformer-Big model}
\begin{figure}[ht]
\centering
  \begin{subfigure}{0.42\textwidth}
  \centering
  \includegraphics[width=1.0\textwidth]{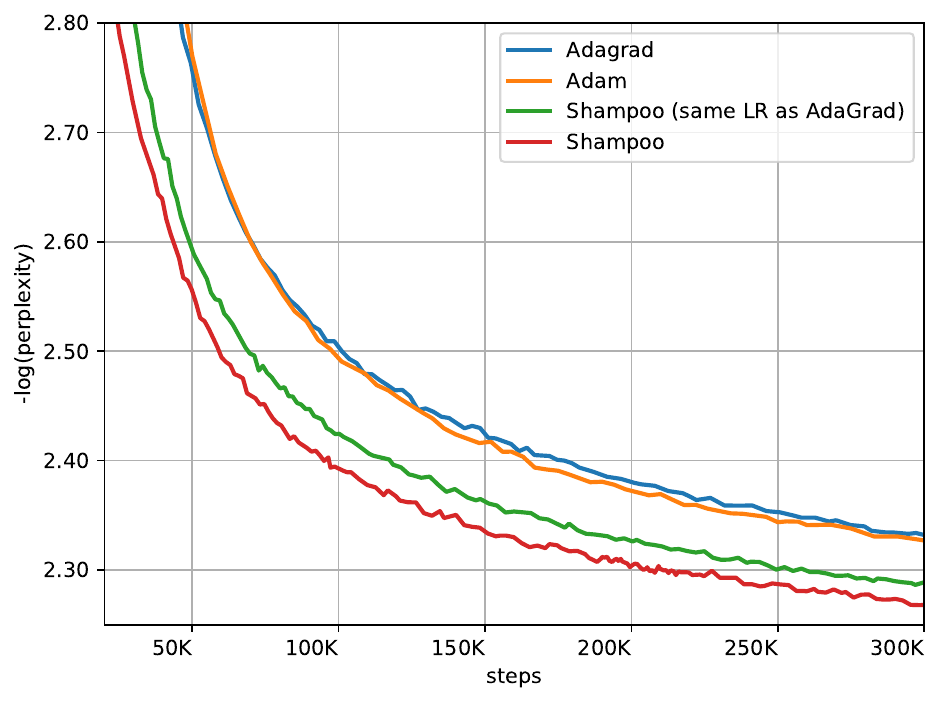}
  \caption{\small{Batch size: 384}}
  \label{fig:en_fr_perplexity_big_same_lr}
  \end{subfigure}
  \begin{subfigure}{0.42\textwidth}
  \centering
  \includegraphics[width=1.0\textwidth]{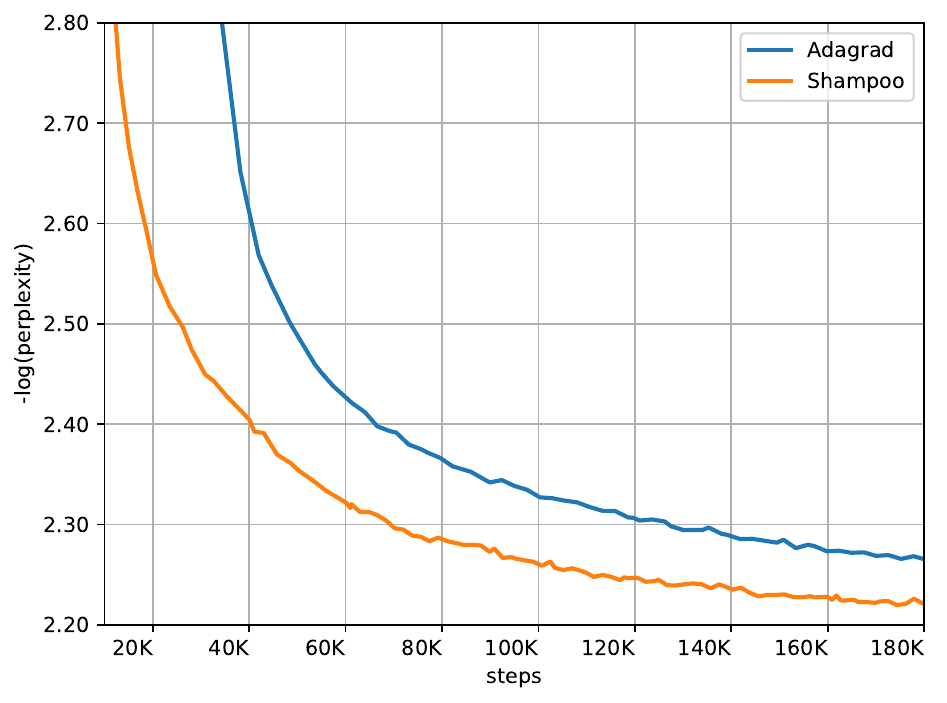}
  \caption{\small{Batch size: 1536}}
  \label{fig:en_fr_perplexity_big_big_batch}
  \end{subfigure}
  \caption{\small{Test log-perplexity of a Transformer-Big model on WMT'14 \entofr. (a) Shampoo converges faster than AdaGrad ($\approx\!2$x faster in steps), and allows larger learning rates; due to the large overhead in step time, this results in only 30\% improvement in wall-time. (b) Larger batch sizes reduce the optimizer overhead from  40\% to 19\%, resulting in an \emph{end-to-end improvement of 41\%} in wall-time for convergence.}}
\end{figure}

We also ran experiments with a larger Transformer model with
375.4M parameters, consisting of 6 layers for its encoder and decoder. Each
layer is composed of 1024 model dimensions, 8192 hidden dimensions, and 16
attention heads. Our results are presented in \cref{fig:en_fr_perplexity_big_same_lr} where
again we see an improvement in the end-to-end wall-clock time. For the softmax,
embedding and the projection fully-connected layer (with 8192 hidden dimensions)
we only make use of the left preconditioner. The step time is dominated
by the preconditioned gradient computation which can be reduced by sub-blocking
the layers.

\paragraph{On the overhead of the optimizer.}
We show the computational and memory complexity of the Shampoo extensions
(described in \cref{subsec:large_tensors}) in
\cref{sec:second-order-experiments}. The overhead from computing the statistics,
as well as from computing the preconditioned update for single step of training,
can be further reduced by increasing the batch size (indeed, these overheads are
independent of the batch size) as shown in
\cref{fig:en_fr_perplexity_big_big_batch} where the overhead dramatically
reduces from 40\% to 19\%.

\subsection{Ads Click-Through Rate (CTR) prediction}
We trained the Deep Learning Recommendations Model (DLRM) of
\citet{naumov2019deep} on the terabyte Criteo click logs dataset for online
advertisement click-through-rate prediction task \citep{labs_2017}. We compared
Shampoo against the highly tuned SOTA baseline from MLPerf v0.7 training
benchmarks \citep{wu2020developing}. We trained the model with a batch size of
65536 for 64000 steps (1 epoch). Here we apply Shampoo to a) hidden layers b) both embedding and hidden layers. We found that Shampoo achieves the target accuracy of 80.25\% in only
30.97K steps compared to 64K steps for the baseline. Moreover, Shampoo achieves
new state-of-the-art performance of 80.56\% AUC (an $\approx\!0.3\%$ improvement) on this dataset, note that an improvement of 0.1\% is considered
significant in this task; see \citealp{rong2020distributed, wang2017deep}. Preconditioning of embedding layers further reduced the number of steps needed to
reach the target accuracy from 39.96K to 30.97K. 

\begin{figure}[h!]
\centering
  \begin{subfigure}{0.42\textwidth}
  \centering
  \includegraphics[width=1.0\linewidth]{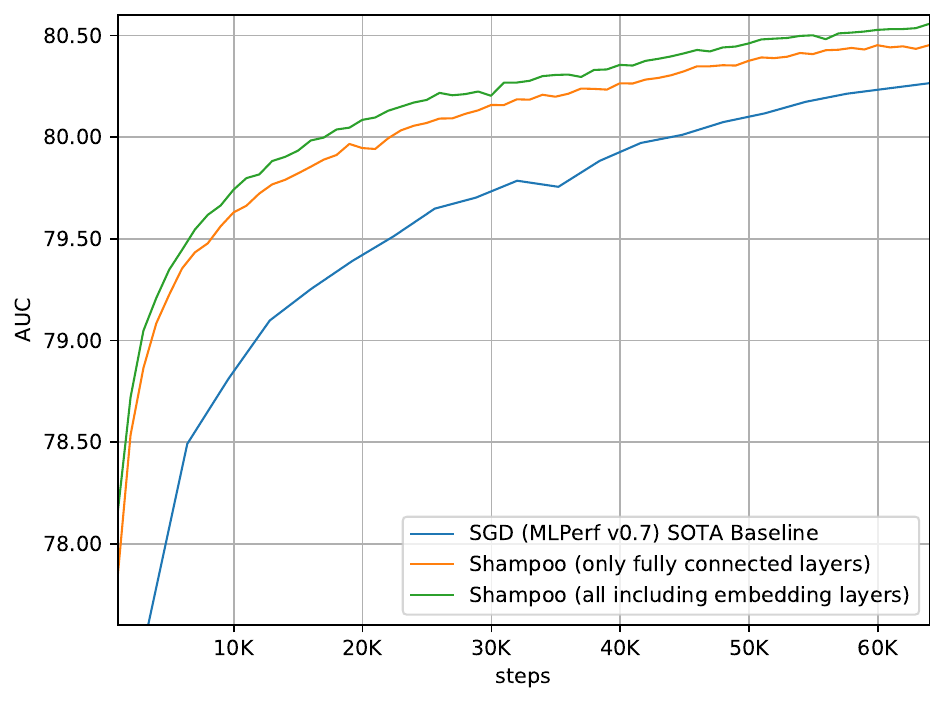}
  \caption{\small{Test AUC on the Criteo-1Tb dataset.
  }}
  \label{fig:criteo}
  \end{subfigure} 
  \begin{subfigure}{0.42\textwidth}
  \centering
  \includegraphics[width=1.0\linewidth]{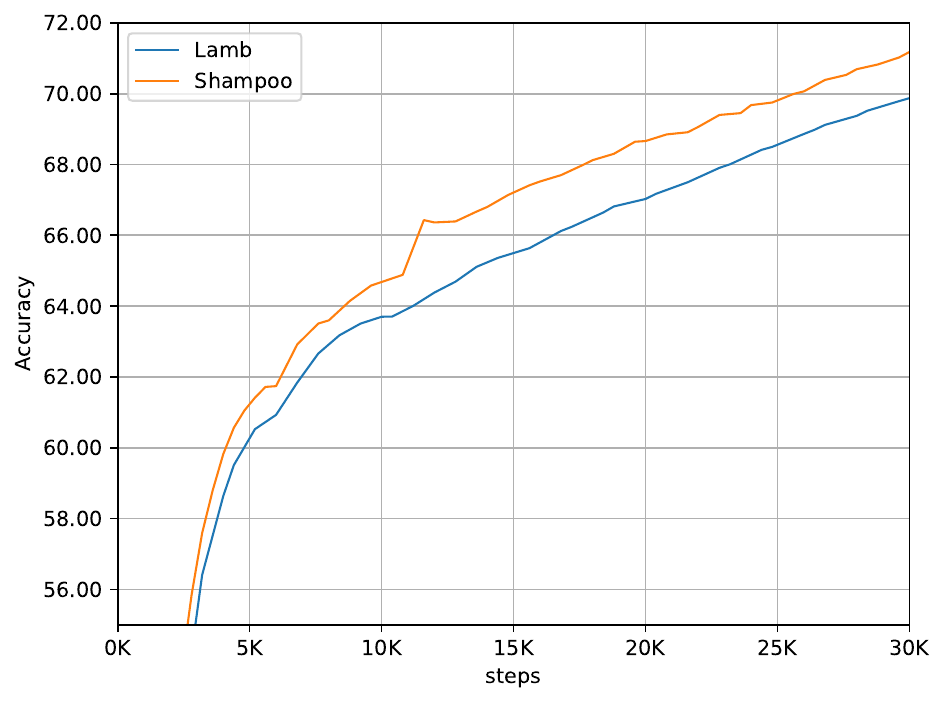}
  \caption{\small{Masked Language accuracy on BERT-Large.
  }}
  \label{fig:bert}
  \end{subfigure}
  \caption{\small (a) Shampoo reaches a target AUC of 80.25\% in half as many steps with preconditioning embedding layers improving the results, and achieves a new state-of-the-art AUC of 80.56\%; (b) Shampoo converges in $\approx\!16\%$ fewer steps, and achieves $\approx\!1\%$ higher MLM accuracy than the baseline on BERT-Large. }
\end{figure}

\subsection{Language modeling}

We trained BERT-Large \citep{devlin2018bert} for the language modeling task on the concatenation of
Wikipedia and BooksCorpus, with 2.5B and 800M words respectively.  BERT-Large is
a bidirectional transformer model containing 24 transformer blocks with
1024 hidden dimensions and 16 self attention heads, a total of 340M parameters.
BERT is set up to jointly optimize two objectives: (a) masked language model
(Masked-LM) loss where the task is to predict masked tokens based on surrounding
context, and (b) next sentence prediction (NSP) loss where the task is to
predict whether two given sentences are consecutive in the text.
In~\cref{fig:bert} we compare our results against the current state of the art
in training BERT~\citep{you2019large}. Models were trained with batch size of 16K, tuning details are in~\cref{sec:StepTime}.

\subsection{Image classification}

We trained a ResNet-50 model \citep{resnet} on the ImageNet-2012
\citep{russakovsky2015imagenet} dataset and compared it against the
state-of-the-art baseline using Nesterov momentum as well as LARS optimizers. We base our experiments JAX baseline available from \cite{mattson2019mlperf} where the target
criteria is reaching 75.9\% accuracy. See results in~\cref{tbl:imagenet}; in
particular, we find that Shampoo reaches the target accuracy in 1729 steps compared to 2512 steps by first order methods. Code as well as tuning details are available in \href{https://bit.ly/3uXXtKy}{https://bit.ly/3uXXtKy}.

\begin{table}[h!]
  \caption{\label{tbl:imagenet}\small Epochs and steps to MLPerf target accuracy of 75.9\% with a ResNet-50.}
  \centering
  \begin{tabular}{p{2.4cm}p{1.75cm}p{1.4cm}p{1.0cm}}
  \toprule
  {\sc Optimizer} & {\sc Batch Size} & {\sc Epochs} & {\sc Steps} \\
  \midrule
  Nesterov &  32768 & 64 & 2512 \\
  \midrule
  LARS & 32768 & 64 & 2512 \\
  \midrule
  \textbf{Shampoo} & \textbf{32768} & \textbf{44}  & \textbf{1729}  \\
  \bottomrule
 \end{tabular}
\end{table}

\section{Concluding Remarks}
\label{sec:concluding-remarks}

We have presented an implementation of a second order optimizer, and
demonstrated step time as well as wall time improvements on multiple large tasks
in different domains---in each case our implementation performed as well or
better than  state-of-the-art optimizers specialized for each domain. 
We hope that this work will influence future hardware accelerator design
and runtime software:
\begin{itemize}
\item Most second order methods use symmetric matrices, but we haven't found
support for typing operands as symmetric, which can reduce flops and storage by
up to $\approx\!50\%$.
\item Several optimizations that are currently tuned towards first order methods
could be extended to second order methods. For example, weight update sharding
pattern matches first order methods \citep{xu2020automatic} and dramatically
reduces the time spent in the update step as well as memory used. This change
can also be applied to Shampoo with blocked preconditioners---but we do not have
support for it yet as it requires compiler level support, and is not expressible
at the program layer. Currently every core must update all layers which is quite
inefficient.
\item Mixed precision algorithms may work for inverse pth roots and can allow more frequent preconditioner computation.
\item Increased memory per chip can allow larger preconditioners.
\item Hardware support for high-precision arithmetic in accelerators can allow more frequent preconditioner computation. The benefits of high precision arithmetic for optimization run counter to the prevailing wisdom in ML%
\footnote{For example, \citet{gupta2015deep} say ``it is well appreciated that in the presence of statistical approximation and estimation errors, high-precision computation in the context of learning is rather unnecessary...'' and \citet{higham2019simulating} say ``machine learning provides much of the impetus for the development of half precision arithmetic in hardware...''.}which has led to the focus on low-precision formats such as bfloat16~\citep{bfloat16google}.
\item Hardware support for storing/packing and using upper/lower triangular matrices efficiently, as available in libraries like LAPACK.
\end{itemize}
Our hope is that these suggestions could result in innovations
that would make second-order methods practical across more domains and models,
especially in data limited regimes where we may not able to amortize the
latency added in the data transfer between the accelerator and the CPU.

\bibliography{bib}
\bibliographystyle{icml2021}

\appendix

\section{Deferred proofs}\label{sec:Proofs}

\begin{proof}[of~\cref{lemma:extend}]
	Lemma~8 in~\cite{shampoo-icml} shows that $\widehat{H}_t \preceq rL_t \otimes I_n$
	and $\widehat{H}_t \preceq rI_m \otimes R_t$. By using Ando's inequality~\citep{ando2004geometric},
	we get
	\begin{align*}
		\widehat{H}_t &\preceq r(L_t \otimes I_n)^{1/p}(I_m \otimes R_t)^{1/q} \\
		    &= r(L_t^{1/p} \otimes I_n)(I_m \otimes R_t^{1/q})\\
				&= rL_t^{1/p} \otimes R_t^{1/q} ~,
	\end{align*}
	which concludes the proof.
\end{proof}

This lemma immediately allows us to prove a regret bound for Shampoo with extended exponents:

\begin{theorem} \label{thm:regret-2d}
Assume that the gradients $G_1,\ldots,G_T$ are matrices of
rank at most $r$.
Then the regret of Shampoo with extended exponents compared to any
$W^\st \in \reals^{m \times n}$ is bounded as follows,
\begin{align*}
  \sum_{t=1}^T f_t(W_t) - \sum_{t=1}^T f_t(W^\st)
  \le
  \sqrt{2r}D \trace(L_T^{\frac{1}{2p}}) \trace(R_T^{\frac{1}{2q}})
  ~,
\end{align*}
where
$$
  L_T = \eps I_m + \sum_{t=1}^T G_t G_t\tr \;,\;\;
  R_T = \eps I_n + \sum_{t=0}^T G_t\tr G_t \;,\;\;
  D = \max_{t \in [T]} \norm{W_t-W^\st}_2 ~.$$
  and $1/p + 1/q = 1, p, q \geq 1$.
\end{theorem}

\begin{proof}
The proof follows the proof of Theorem 7 in~\cite{shampoo-icml}. Let $H_t =
L_t^{\frac{1}{2p}} \otimes R_t^{\frac{1}{2q}}$. Then the update rule of the
extended Shampoo algorithm is equivalent to $w_{t+1} = w_t - \eta H_t^{-1} g_t$.
Since $0 \preceq L_1 \preceq \ldots \preceq L_T$ and $0 \preceq R_1 \preceq
\ldots \preceq R_T$, standard properties of the Kronecker product and the
operator monotonicity of the function $x \mapsto x^\alpha$ for $\alpha \leq 1$
(an immediate consequence of Ando's inequality) ensure that $0 \preceq H_1
\preceq \ldots \preceq H_T$.

Following the aforementioned proof, we have the regret bound 
$$
  \sum_{t=1}^T f_t(W_t) - \sum_{t=1}^T f_t(W^\st) 
  \leq 
  \frac{D^2}{2\eta} \trace(H_T) + \frac{\eta}{2} \sum_{t=1}^T \norm{g_t}^2_{H_t^*},
$$ 
where $D = \max_t \norm{W_t - W^\st}_2$. Define $g_t = \vec(G_t)$ and
$\widehat{H}_t = (\eps I_m + \sum_{s=1}^t g_s g_s\tr)^{1/2}$,
then~\cref{lemma:extend} shows that $\widehat{H}_t \preceq \sqrt{r}H_t$, using
operator monotonicity. Using this equation twice, along with Equation (6) from
the proof of Theorem 7, we have
$$
  \sum_{t=1}^T \norm{g_t}^2_{H_t^*} 
  \leq 
  \sqrt{r} \sum_{t=1}^T \norm{g_t}^2_{\widehat{H}_t^*} 
  \leq 
  2 \sqrt{r} \trace(\hat{H}_T) \leq 2r \trace(H_T)
  .
$$ 
This gives us 
$$
  \sum_{t=1}^T f_t(W_t) - \sum_{t=1}^T f_t(W^\st) 
  \leq 
  \frac{D^2}{2\eta} \trace(H_T) + \eta r \trace(H_T)
  .
$$
Setting $\eta = D/\sqrt{2r}$ and observing that $\trace(H_t) =
\trace(L_t^{1/2p}) \trace(R_t^{1/2q})$ gives us the required bound.
\end{proof}

\begin{proof}[of~\cref{lemma:extendblock}]
Let $x \in \reals^{mk}$, and $x = [x_1, x_2, \ldots, x_k]$, where $x_j \in
\reals^{m}$. Then 
\begin{align*}
  x\tr \widehat{H}_t x &= \epsilon\norm{x}_2^2 + \sum_{s=1}^t x\tr g_s g_s\tr x
  = \epsilon\norm{x}_2^2 + \sum_{s=1}^t (g_s\tr x)^2
  = \epsilon\norm{x}_2^2 + \sum_{s=1}^t \biggl(\sum_{j=1}^k g_{s,j}\tr x_j\biggr)^2 \\
  &\leq k\epsilon\norm{x}_2^2 +  k\sum_{s=1}^t \sum_{j=1}^k ( g_{s,j}\tr x_j)^2
  = k \sum_{j=1}^k \biggl(\epsilon\norm{x_j}_2^2 + \sum_{s=1}^t x_j\tr g_{s,j}^{} g_{s,j}\tr x_j\biggr) \\
  &= k \sum_{j=1}^k x_j\tr \bigl(\epsilon I_m +  \sum_{s=1}^t g_{s,j}^{} g_{s,j}\tr\bigr) x_j 
  = k \sum_{j=1}^k x_j\tr B_t^{(j)} x_j = k x\tr B_t x
  .
\end{align*}
Here we used the inequality $\bigl(\sum_{j=1}^k \alpha_j\bigr)^2 \leq k
\sum_{j=1}^k \alpha_j^2$, which follows from the convexity of $x \mapsto x^2$
(or from the fact that variance of a random variable is non-negative).
\end{proof}

This lemma once again allows us to prove a regret bound, exactly following the proof of the regret bound above:

\begin{theorem} \label{thm:regret-block}
Assume that the gradients are $g_1,\ldots, g_T \in \reals^{mk}$, and let $g_i =
[g_{i,1}, \ldots, g_{i,k}]$ where $g_{i,j} \in \reals^m$. Then the regret of
Shampoo with blocking compared to any $w^\st \in \reals^{mk}$ is bounded as
follows:
\begin{align*}
\sum_{t=1}^T f_t(w_t) - \sum_{t=1}^T f_t(w^\st)
\le
\sqrt{2k}D \sum_{j=1}^k \trace\biggl(\biggl(\eps I_m + \sum_{t=1}^T g_{t,j}^{} g_{t, j}\tr \biggr)^{\frac{1}{2}}\biggr)
~.
\end{align*}
\end{theorem}

The two regret bounds can be combined to show that Shampoo with both extensions
also converges.

\section{Comparison with K-FAC}\label{sec:KFAC}

K-FAC is a natural gradient algorithm, and approximates the curvature of the loss using the Fisher Information Matrix:
\begin{align*}
    {\mathbf F} 
    =  \mathop{\mathbb{E}}_{p(x \vert \theta)} \left[ \nabla \log p(x \vert \theta) \, \nabla \log p(x \vert \theta)\tr \right]
    =  \mathop{\mathbb{E}}_{p(x \vert \theta)} \left[ g_{p(x \vert \theta)} \, g_{p(x \vert \theta)}\tr \right]
    .
\end{align*}
For a fully connected layer with $W \in \reals^{m\times n}$, where $Wx = s$, the
gradient for the layer $G_t\in\reals^{m\times{}n}$ can be written via the chain
rule as $G_t = \nabla_s \ell(s_t,y_t) x\tr$ and in vectorized form as: $
\nabla_s \ell(s_t,y_t) \otimes x $.
We can then write the Fisher information matrix as:
\begin{align*}
   {\mathbf F} &= \mathop{\mathbb{E}}_{p(x \vert \theta)} \left[ (\nabla_s \ell(s_t,y_t) \otimes x ) \, ( \nabla_s \ell(s_t,y_t) \otimes x )\tr \right]\\
    &=  \mathop{\mathbb{E}}_{p(x \vert \theta)} \left[ (\nabla_s \ell(s_t,y_t)  \nabla_s \ell(s_t,y_t)\tr ) \otimes \, ( x_t x_t\tr) \right]
    .
\end{align*}
Assuming independence between $ \nabla_s \ell(s_t,y_t) $ and $x$, K-FAC rewrites the Fisher in tractable form as:
\begin{align*}
    {\mathbf F}
    &\approx \mathop{\mathbb{E}} \left[ (\nabla_s \ell(s_t,y_t)  \nabla_s \ell(s_t,y_t)\tr ) \right] \otimes \, \mathop{\mathbb{E}} \left[x_t x_t\tr \right] 
    .
\end{align*}
If we let $D = \mathop{\mathbb{E}} \left[ (\nabla_s \ell(s_t,y_t)  \nabla_s
\ell(s_t,y_t)\tr ) \right]$ and $X = \mathop{\mathbb{E}} \left[x_t x_t\tr
\right]$, the update rule then becomes:
\begin{align*}
  W_{t+1} 
  &\approx 
  W_t - \eta\, D^{-1} G_t X^{-1} 
  .
\end{align*}

We note some of the differences and similarities between the two updates here.
KFAC preconditioners use exponent of $-1$ (as original Fisher is inverted) whereas 
Shampoo uses $-1/2p$ where $p$ is the rank of the tensor.  KFAC computes
statistics based on gradients with labels sampled from the model's predictive
distribution (hence requiring strictly more computation) where as Shampoo relies
on the gradient of the mini-batch. 

Now we can compute each term in the Shampoo preconditioners as:
\begin{align*}
    G_t G_t\tr &= \nabla_s \ell(s_t,y_t) x_t\tr x_t \nabla_s \ell(s_t,y_t)\tr 
    = \norm{x_t}_2^2 \nabla_s \ell(s_t,y_t)\nabla_s \ell(s_t,y_t)\tr
    ;
    \\
    G_t\tr G_t &=  x_t \nabla_s \ell(s_t,y_t)\tr \nabla_s \ell(s_t,y_t) x_t\tr
    = \norm{\nabla_s \ell(s_t,y_t)}_2^2 x_t x_t\tr
    .
\end{align*}
Dividing by the scale, and taking expectations on both sides:
\begin{align*}
    \mathop{\mathbb{E}} \left[ \frac{G_t G_t\tr}{\norm{x_t}_2^2}\right] &=  \mathop{\mathbb{E}} \left[\nabla_s \ell(s_t,y_t)\nabla_s \ell(s_t,y_t)\tr\right] = D
    ;
    \\
    \mathop{\mathbb{E}} \left[ \frac{G_t\tr G_t}{\norm{\nabla_s \ell(s_t,y_t)}_2^2}\right]  &=  \mathop{\mathbb{E}} \left[x_t x_t\tr\right] = X
    .
\end{align*}
This shows that K-FAC preconditioners are closely related to Shampoo preconditioners,
 especially when one uses the empirical Fisher \citep{kunstner2019limitations}.

The main difficulty in implementing K-FAC on a model is that current optimizer
APIs make it difficult to send additional information such as $\norm{x_t}_2^2,
\norm{\nabla_s \ell(s_t,y_t)}_2^2$ to the optimizer, so K-FAC implementations
have to register the structure of each layer. 
Moreover, due to the dependence of K-FAC on the structure of the network, it is
difficult to implement standard operators like batch norm, weight norm, layer
norm, etc., which are prevalent in the tasks and models we considered.
For example, if we write a fully connected layer with weight norm as $s =
{Wx}/{\norm{W}}$, then the gradient $$G_t = \frac{1}{\norm{W}} \nabla_s
\ell(s_t,y_t) x\tr - \frac{\nabla_s \ell(s_t,y_t)\tr Wx}{\norm{W}^3}W,$$ so
rewriting $\E[\vec(G_t)\vec(G_t)\tr]$ as a Kronecker product is not an easy
task.

The similarity between K-FAC and Shampoo preconditioners also allows us to use techniques explored by the K-FAC community for Shampoo. One of the extensions for KFAC is the E-KFAC algorithm~\citep{george2018fast} which constructs a better approximation of the Fisher matrix by using the eigenbasis computed from the Kronecker approximation, but rescaling the eigenvalues to match the diagonal of the Fisher matrix in this eigenbasis. This method produces a provably better approximation, and can immediately be applied to Shampoo too with a simple modification: 

Let $\hat{H}_t \approx L_t^{\sfrac{1}{2}} \otimes R_t^{\sfrac{1}{2}}$. Let the singular value decompositions of the factors be $L_t^{\sfrac{1}{2}}  = UDU\tr$ and $R_t^{\sfrac{1}{2}} = VD'V\tr$. Then $L_t^{\sfrac{1}{2}} \otimes R_t^{\sfrac{1}{2}} = (U \otimes V) (D \otimes D') (U\otimes V)\tr$. Now the EKFAC correction replaces $D \otimes D'$ by the optimal diagonal 
\begin{align*}
  \Lambda &= \diag((U \otimes V)\tr \hat{H}_t (U \otimes V))\\
  &= \epsilon I + \sum_{s=1}^t \diag((U \otimes V)\tr \vec(G_s)\vec(G_s)\tr(U \otimes V))\\
  &= \epsilon I + \sum_{s=1}^t \diag(\vec(U\tr G_s V)\vec(U\tr G_s V)\tr)\\
  &=  \epsilon I +  \sum_{s=1}^t \vec(U\tr G_s V)^{\odot 2}
  ,
\end{align*}
Thus we can approximately compute $\Lambda_{t+1} \approx \Lambda_t + (U\tr G_t V)^{\odot 2}$, and the new update becomes: $W_{t+1} = W_t - \eta_t U( \Lambda_t^{-\sfrac{1}{2}} \bullet (U\tr G_t V)) V\tr$. This technique does have the disadvantage that it requires computing the singular value decompositions (which we already observed are much slower than coupled Newton iterations), and doubles the number of matrix multiplications in the preconditioned gradient computation. At this time our experiments did not show significant improvements over the standard Shampoo implementation, but we plan to explore this further.

\section{Shampoo for embedding layers}\label{sec:Embedding}
In modern networks, embedding layers are usually very large, and even computing the left preconditioner as described in \cref{subsec:large_tensors} can be prohibitively expensive. However we can take advantage of the fact that the inputs to the network are very sparse, and use this to reduce the computation significantly.

Let our input example to such a network consist of a set of categorical
features: each feature such as user language, user country etc consists of one
out of a set of options. Then the output of the embedding layer is the
concatenation of the embeddings for each such feature. If the embeddings are of
width $d$ and there are $N$ such embeddings, then the embedding layer is $W \in
\reals^{d \times N}$. The input can be represented as $x \in \reals^{N \times
m}$, where $m$ is the number of categorical features, and each column is
one-hot: if the $k$-th feature is $x(k)$, then $x_{jk} = \delta_{j, x(k)}$. The
output of the layer is $y = Wx$.

Now $G =\nabla_W \ell = \nabla_y \ell \, x\tr$, so $GG\tr = \nabla_y \ell\, x\tr
x\, \nabla_y\ell\tr$. But $x\tr x = {\bf I}_m$, so $GG\tr = \nabla_y \ell\,
\nabla_y \ell\tr$. Thus we can compute the preconditioner for $W$ by computing
it on the output of the embedding layer, and this is a much smaller computation
since $y$ is of dimension $b \times m$, this computation is $O(d^2m)$ rather
than $O(d^2N)$. Note that sparse multiplication would also be $O(d^2 m)$, but
accelerators usually implement sparse operations by densifying the tensors.

If each column of $x$ is multi-hot, as is the case when the features are words and their embeddings are averaged, $x\tr x$ is a diagonal matrix, where each diagonal entry is a function of the number of ones in each column of $x$. Computing $GG\tr = \nabla_y \ell (x\tr x) \nabla_y \ell\tr$ is still $O(d^2m)  \ll O(d^2 N)$.

\section{A coupled Newton iteration for computation of inverse {\it p}-th roots}
\label{sec:SNopt}
The Newton method for solving the matrix equation $X^{-p} - A = 0$ produces the iteration $X_{k+1} = \frac{1}{p}[(p+1)X_k - X_k^{p+1} A]$, where we take $X_0 = \frac{1}{c}I$. This iteration satisfies $X_k \rightarrow A^{-1/p}$ as $k \to \infty$, but it is not numerically stable. Introducing the matrix $M_k = X_k^p A$, we get 
$$X_{k+1} = X_k\biggl(\frac{(p+1)I - M_k}{p}\biggr), \qquad X_0 =
\frac{1}{c}I,$$ and $$M_{k+1} = X_{k+1}^p A = \biggl(\frac{(p+1)I -
M_k}{p}\biggr)^p X_k^p A = \biggl(\frac{(p+1)I - M_k}{p}\biggr)^p M_k,  \qquad
M_0 = \frac{1}{c^p}A,$$ since $X_k, M_k$ and $A$ commute with each other. This
is the coupled Newton iteration for computing inverse $p$-th
roots, and was shown to be numerically stable in~\citep{guo2006schur,iannazzo2006newton}.

We implemented the following optimizations to the coupled Newton iteration method:
\begin{itemize}
\item {\it Warm Start}: The coupled Newton iteration to compute $G^{-1/p}$
starts with $X = I, M = G$ and maintains the invariant $M = X^pG$ while driving
$M \rightarrow I$, resulting in $X \rightarrow G^{-1/p}$. We need to find the
$p$-th root of a sequence $G_t$, so we instead set $X =
G_{\smash{t}}^{\smash{-1/p}}, M = X^p G_{t+1}$; since the difference between
$G_t$ and $G_{t+1}$ is small, this ensures that $M$ is already close to $I$. In
our experiments warmstart improves convergence (by upto 4x fewer steps), in some cases.
Note that for the coupled iteration to work, it is necessary that $XM = MX$ --- this is only
approximately true if we initialize $X =
G_{\smash{t}}^{\smash{-1/p}}, M = X^p G_{t+1}$, so we monitor the commutator $[X, M] = XM - MX$, 
and if it diverges, we abort the warm start and re-initialize $X = I, M = G_{t+1}$.
\item {\it Scaled damping}. In order to avoid numerical problems, as well as addressing 
rank deficient matrices, we add a damping factor before taking the inverse root: $G + \epsilon_d I$.
In our experiments we discovered that scaling the $\epsilon_d$ by the spectral norm of $G$ (the
largest eigenvalue, since are matrices are positive semi-definite) improves the performance of
the optimizer --- intuitively, we scale the damping to match the scale of the matrix.

\end{itemize}

\begin{algorithm}[htb]
\begin{algorithmic}[1]
\Procedure{MaxSV}{${\bf G}$}
   \State {\bf Parameters}: $\epsilon > 0$, $n_{\text{step}}$
   \State ${\bf v} \in \reals^n$, where ${\bf G} \in \reals^{n\times n}$
   \State $i = 0,\, \mbox{error} = \infty,\, \lambda = 0$
   \While{$i < n_{\text{step}}$ and $\mbox{error} > \epsilon$}
   	\State $\hat{\bf v} = \sfrac{\bf v}{\Vert \bf v \Vert}$
	\State ${\bf v} = {\bf G\hat{v}}$
	\State $\lambda_{\text{old}} = \lambda; \lambda = {\hat{\bf{v}}}\tr{\bf v}$
	\State $\mbox{\text{error}} = |\lambda - \lambda_{\text{old}}|; i = i + 1$
   \EndWhile
   \State  \textbf{return}  $ \lambda$%
\EndProcedure
\State
\Procedure{CoupledIteration}{${\bf G}$, $p \in \mathbb{N}$, ${\bf X}$ (optional)}%
  \State {\bf Parameters:} $\epsilon > 0$, $\epsilon_d > 0$%
  \State {\bf Outputs: ${\bf G}^{-\sfrac{1}{p}}$}
  \State $\lambda_{\mbox{max}} = $ {\sc MaxSV}($G$)
  \State ${\bf G} = {\bf G} + \epsilon * \lambda_{\mbox{max}} * I$
  \State $\alpha = -\frac{1}{p}$
  \If{${\bf X}$ is provided}
    \State ${\bf M} = {\bf X}^p {\bf G}$
  \Else
    \State $z = \frac{1 + p}{2\Vert {\bf G}\Vert_F}$
    \State ${\bf X} = \frac{1}{z^\alpha} {\bf I}$
    \State ${\bf M} = z{\bf G}$
  \EndIf
  \While{$\Vert {\bf M} - {\bf I} \Vert_\infty > \epsilon$}
        \State ${\bf M}_1 = (1 - \alpha) {\bf I} + \alpha {\bf M}$
  	\State ${\bf X} =  {\bf X}  {\bf M}_1$
	\State ${\bf M} =  {\bf M}_1^p {\bf M}$
  \EndWhile
  \State \textbf{return} ${\bf X}$
\EndProcedure
\end{algorithmic}
\caption{A coupled Newton iteration procedure for computing inverse $p$-th roots of a PSD matrix, with warm start and singular value projection}
\label{pth_root_algorithm}
\end{algorithm}

\begin{figure}[htb]
  \centering
  \includegraphics[width=0.47\textwidth]{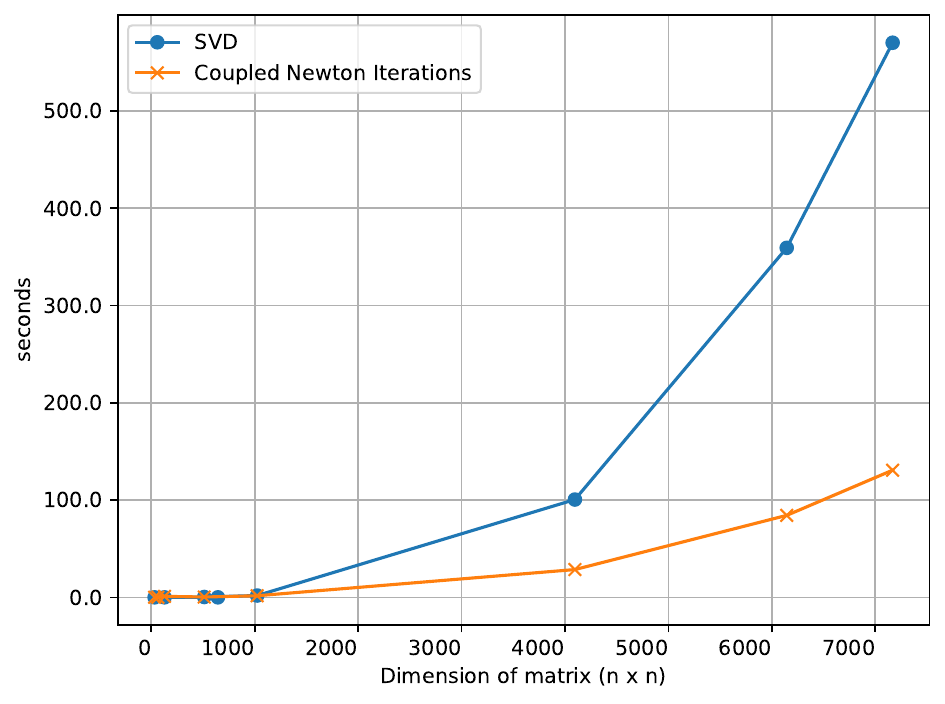}
  \hfill
  \includegraphics[width=0.47\textwidth]{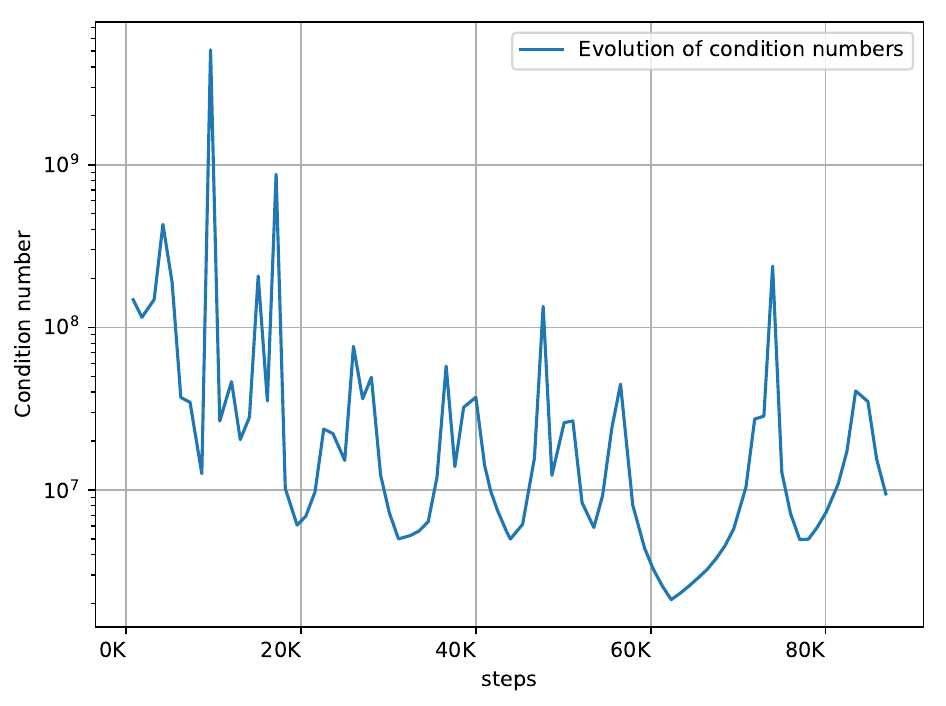}
  \caption{\small{Benchmarks on computing inverse-pth root for statistics of
  varying dimensions (left), and the condition numbers for $L_t$ of a layer in
  the transformer model over time (right). We find that the coupled Newton
  iteration method can effectively utilize the CPUs and give large walltime
  improvements compared to SVD (that relies on bidiagonal divide-and-conquer).
  These were measured without warmstart which provides additional speedup of
  upto 4x by reducing the number of iterations to the solution.These were
  measured on Intel Skylake CPUs. Note that since $\approx \log_2 (\frac{1}{p} \kappa(L_t))$
  bits of precision are lost in computing $p$-th roots, 64-bit arithmetic becomes necessary.}}
  \label{fig:condition_numbers}
\end{figure}

\section{Implementation details of Shampoo}

Our implementation of the Shampoo algorithm for fully-connected layers is described in \cref{shampoo_algorithm}. 
The algorithm can use heavy-ball momentum for its updates, as well an exponential moving average over the preconditioners, like Adam. 
The configuration parameter $\tau_1$ denotes the number of steps between subsequent fetches of the latest available preconditioner by the accelerator. 
$\tau_1$ must be set sufficiently high so that there is enough time for the CPU to complete the computation of the preconditioner asynchronously and pipeline it efficiently, but otherwise its setting does not have a significant effect on convergence. The configuration parameter $\tau_2$ (default value $=1$) determines the frequency of gathering gradient statistics - we update $L_t, R_t$ every $\tau_2$ steps only for efficiency.

\begin{algorithm}[htb]
\begin{algorithmic}[1]
  \State {\bf parameters:} learning rate $\eta_t$, momentum: $\beta_1$, $\beta_2$
        \For{$t = 1, \ldots, T$ }
                \State Receive stochastic gradients $G_t$ for each layer
                \If{$t~\%~\tau_2 = 0$}
                \If{$\beta_2 < 1$}
                \State $L_t \leftarrow \beta_2~L_{t-\tau_2}$ + $\left(1- \beta_2\right)~G_t G_t\tr$
                \State $R_t \leftarrow \beta_2~R_{t-\tau_2}$ + $\left(1 - \beta_2\right)~G_t\tr G_t$
                \Else
                \State $L_t \leftarrow  L_{t-\tau_2}$ + $G_t G_t\tr$
                \State $R_t \leftarrow  R_{t-\tau_2}$ + $G_t\tr G_t$
               \EndIf    
             \EndIf
            \State $D_t \leftarrow  D_{t-1} + G_t \bullet G_t$
            \State $M_t \leftarrow \beta_1~M_{t-1} + \left(1 - \beta_1\right)~D_t^{\odot-1/2} \bullet G_t $
            \If{$t~\%~\tau_1 = 0$}
            \State Gather preconditioners $ L_{\left(t-\tau_1\right)}^{-1/4},R_{\left(t-\tau_1\right)}^{-1/4}$ from CPUs
            \State Send $ L_t,R_t$ to CPU host to compute $L_t^{-1/4}, R_t^{-1/4}$
            \EndIf

            \If{$t > \tau_1$}
            \State $P_t \leftarrow  \beta_1 P_{t-1} + \left(1-\beta_1\right)~L_t^{-1/4} G_t R_t^{-1/4}$
            \State $\eta_t \leftarrow  \eta_0 \Lrnorm{M_t}_F \big/ \Lrnorm{P_t}_F$	        
            \State $W_t = W_{t-1} - \eta_t P_t$           
            \Else
            \State $\eta_t \leftarrow  \eta_0$    	            			
            \State $W_t = W_{t-1} - \eta_t M_t$           
           \EndIf
        \EndFor
\end{algorithmic}
\caption{Sketch of the Shampoo algorithm}
\label{shampoo_algorithm}
\end{algorithm}

\subsection{Computation cost of Shampoo}

We capture the computational and memory complexity under various schemes
described  in \cref{subsec:large_tensors} of handling large layers in
\cref{complexity}.

\begin{table*}
\centering
 \begin{tabular}{l c c} 
 \hline
 \sc Type & \sc Computation & \sc Memory  \\ [0.5ex] 
 \hline
All preconditioner $W_t$: $[n, m]$ & $O(n^{2}m + m^{2}n)$ & $ O(n^{2} + m^{2}) $ \\ 
 \hline
 Left only preconditioner for $W_t$: $[n, m]$  & $O(n^{2}m)$ & $ O(n^{2})$ \\ 
 \hline
 Preconditioner: block size $b$ & $O(mnb)$ &  $ O(mn) $ \\ 
 \hline
\end{tabular}
\caption{Computational and memory complexity of variants of Shampoo.}
\label{complexity}
\end{table*}

\section{Experimental comparison with second order optimizers}
\label{sec:second-order-experiments}

In ~\cref{fig:autoencoders}, we showed the results of Shampoo against K-BFGS and KFAC on standard autoencoder 
problems: MNIST\footnote{Downloadable at \url{yann.lecun.com/exdb/mnist/}}, 
FACES\footnote{Downloadable at \url{www.cs.toronto.edu/~jmartens/newfaces_rot_single.mat}} and 
CURVES\footnote{Downloadable at \url{www.cs.toronto.edu/~jmartens/digs3pts_1.mat}}.
We used code from the git repository released
with~\citep{goldfarb2020practical} at \url{github.com/renyiryry/kbfgs_neurips2020_public}, 
and used the hyperparameters they found to be optimal 
for each of these algorithms for each dataset. We tuned Shampoo by hand, and found that the
parameter settings described below gave reasonable results --- the main observation we make from
this experiment is that with appropriate tuning all these algorithms can perform well.

\begin{table*}[htb]
\centering
 \begin{tabular}{l c c c c} 
 \hline
 \sc Task & \sc Learning Rate & \sc Ridge Epsilon & \sc  Momentum & \sc Warmup\\ [0.5ex] 
 \hline
MNIST & 0.032 & $10^{-3}$ & 0.9 & 0\\
\hline
FACES & 0.033 & $5 \times 10^{-6} $ & 0.9 & 0.99\\
 \hline
 CURVES & 0.1 & $3.5 \times 10^{-6}$ & 0.99 & 0.999\\
 \hline
\end{tabular}
\caption{Hyperparameters used by Shampoo for autoencoders.}
\label{}
\end{table*}

The standard update for Shampoo described in~\cref{sec:background} is $W_{t+1} = W_t - \eta L_t^{-1/4} G_t R_t^{-1/4}$. 
However during our experiments we realized that sometimes we get better results by treating the exponent as a hyperparameter, thus using the update $W_{t+1} = W_t - \eta L_t^{-\frac{\alpha}{2}} G_t R_t^{-\frac{\alpha}{2}}$ for $\alpha \in [0, 1]$. In the above experiments, we used $\alpha = 1$, which corresponds to a Kronecker approximation of the Online Newton Step algorithm~\citep{hazan2007logarithmic}. Furthermore, as described in~\cref{sec:exper-details}, the learning rates for Shampoo were derived from SGD.

\section{Further details on experiments}
\label{sec:exper-details}

\paragraph{Layer wise learning rates.}

\begin{figure}[htb]
\begin{center}
\includegraphics[width=0.5\textwidth]{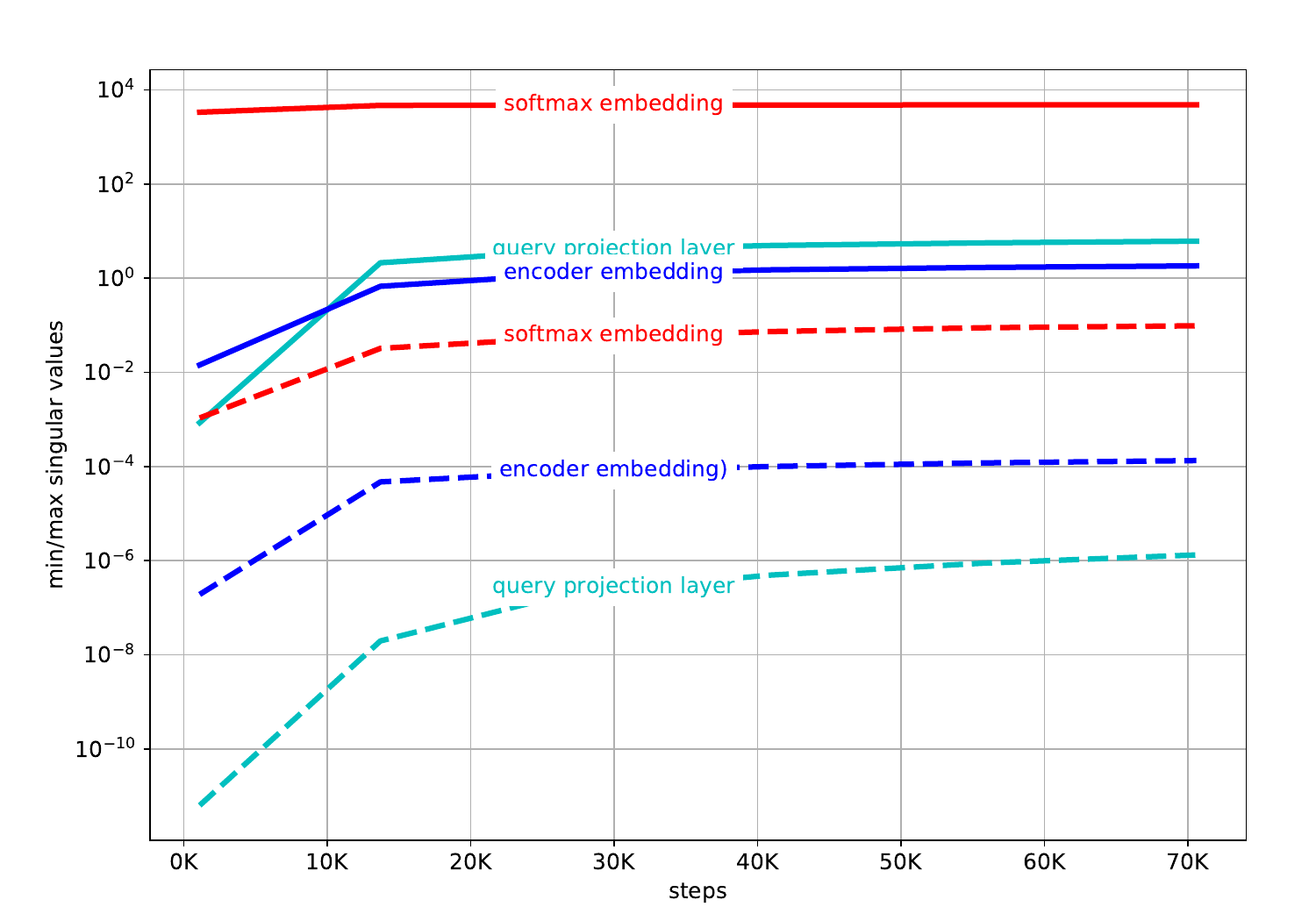}
\end{center}
\caption{Minimum (dashed) and maximum (solid) singular values for statistics matrices of the embedding, softmax and intermediate attention query projection layers.} 
\label{fig:singularvalues}
\end{figure}

As seen in \cref{fig:singularvalues} the step size scale for each layer is
dependent on the operator norm of the preconditioners (inverse-pth root of the
smallest singular value of the statistics matrix) has large spread in its range
which results in optimization instabilities in practice. Moreover, as statistics
as well as preconditioner computation are amortized across many steps the norm
does not grow at every step. Hence, we rely on a learning rate schedule based on
the update directions of a well tuned first order optimizer (in our experiments
we use diagonal AdaGrad for Transformers in machine translation, as well as
Criteo, layer-wise scaling heuristic proposed in LARS/LAMB optimizer, where each
layer's learning rate is set to be ${\Lrnorm{W_t}_F}\big/{\Lrnorm{G_t}_F}$ for
BERT and ResNet training. For example, when used with diagonal AdaGrad: Shampoo
is used to determine the direction of the update, and AdaGrad to determine its
magnitude. 

This procedure termed Grafting in \citep{agarwal2020disentangling} allows us to
bootstrap a reasonable learning rate schedule for a specific problem that is
well tuned, and study the effect of preconditioned gradient directions in
isolation. The weight matrix $W_t$ is updated as $W_t = W_{t-1} - A_t {\bf
\hat{S}_t}$, where:
\begin{align*}
  D_t &= \sum_{s=1}^t G_s \bullet G_s ;\ \  A_t = \eta_0 \, \Lrnorm{D_t^{\odot-1/2} \bullet G_t}_F 
  &\mbox{(Adagrad magnitude)}
	\\
	{\bf \hat{S}_t} &= 
    	\frac{L_t^{-1/4} G_t R_t^{-1/4}
 	}{
    	\Lrnorm{L_t^{-1/4} G_t R_t^{-1/4}}_F
      }  
  &\mbox{(Shampoo direction)}
  .
\end{align*}

\begin{table*}[htb]
\centering
\begin{small}
\begin{tabular}{llclc}
\toprule
\head{3.33cm}{Experiment (TPU cores)} & \head{0.35cm}{Optimizer}& \head{0.5cm}{Batch} & \head{4.0cm}{Optimizer Parameters} & \head{0.5cm}{Warmup}    \\
\midrule
\multirow{3}{*}{Transformer (32)}  &  Adam& 1536 & $\eta=0.000225$, $\beta_{1}=0.9$, $\beta_{2}=0.98$ & 40k steps\\
&  Adagrad& 1536 & $\eta=0.125$, $\beta_{1}=0.95$  & 40k steps\\
&  Shampoo& 1536 & $\eta=0.225$, $\beta_{1}=0.95$, $\kappa=500$ & 40k steps\\
&  &  &  $\tau_1=1000$, $\tau_2=1$  & \\

\midrule
\multirow{3}{*}{Transformer-Big (32)}  &  Adam& 384 & $\eta=0.000154$, $\beta_{1}=0.9$, $\beta_{2}=0.999$ & 40k steps\\
&  Adagrad& 384 & $\eta=0.03$, $\beta_{1}=0.9$  & 40k steps\\
&  Shampoo& 384 & $\eta=0.06$, $\beta_{1}=0.9$, $\kappa=500$ & 40k steps\\
&  &  &  $\tau_1=1000$, $\tau_2=1$  & \\
\midrule
\multirow{2}{*}{Transformer-Big (32)}  &  Adagrad& 1536 & $\eta=0.06$, $\beta_{1}=0.9$ & 40k steps\\
&  Shampoo& 1536 & $\eta=0.08$, $\beta_{1}=0.9$, $\kappa=500$ & 40k steps\\
&  &  & $\tau_1=1000$, $\tau_2=1$  & \\

\midrule
\multirow{4}{*}{Bert-Large (256)}  &  LAMB  & 16384 & $\eta=0.0060$ $\beta_1=0.9$, $\beta_2=0.999$ & 6.4k steps\\
					   &  Shampoo & 16384 &$\eta=0.0060$  $\beta_1=0.9$, $\beta_2=0.999$, & 6.4k steps\\
					   &   &  &  $\lambda_2=10^{-2}$, $\tau_1=400$, $\tau_2=10$ & \\
					   &   &  & Block size: 1024 & \\
\midrule
\multirow{5}{*}{DLRM (32)}  &  SGD  & 65536 & $\eta=0.1$, poly decay(p=2) at 38k steps  &  2k steps\\
					   &  Shampoo & 65536 & $\eta=0.1$  poly decay(p=2) at 38k steps & \multirow{2}{*}{2k steps} \\
				& &  &  $\beta_1=0.9$, $\tau_1=999$, $\tau_2=10$ \\ 
					   & (w/ embd) & 65536 &$\eta_{\text{embd}}=0.31$  & \\
\bottomrule
\end{tabular}
\end{small}

\caption{Experimentation setup, including number of TPU cores, as well hyper-parameters used in our experiments.}
\label{tbl:hparams}
\end{table*}

\subsection{Transformer model on WMT'14 \entofr}

For all optimizers, we make use of a warmup schedule where the learning rate is
increased from 0.0 to $\eta$ over 40k steps. For the smaller transformer
experiments, we use a quadratic warmup, and for the larger transformer
experiments we use a linear warmup. We found that quadratic warmup improves all
optimizers equally and provides a better log-perplexity. For the Adam optimizer
experiments, we use a learning rate decay schedule of the form $\eta_t = \eta
\sqrt{d/t}$, following the suggestion of \citet{vaswani2017attention}.  For the
smaller Transformer experiments, we  tuned the hyperparameters for each
algorithm over 100 trials. We took the best settings for the momentum and
second-moment parameters, and tuned the learning rates until either the model
became unstable, or did not increase performance. For Shampoo, we used a per
layer learning rate derived from AdaGrad (see \cref{sec:exper-details} for
details), and found that for the exact same hyperparameter settings as AdaGrad,
Shampoo provides a modest improvement in performance. Moreover, Shampoo allows
for larger learning rates than AdaGrad does, as shown in
\cref{fig:en_fr_perplexity_big_same_lr}.

\subsection{BERT-Large}\label{sec:StepTime}
Our current implementation showed a 14\% increase in step time for BERT-Large,
nearly wiping out all the gains from reduced number of steps (16\%). We note
that due amount of resources it would require to tune BERT, we used Shampoo with
exact same hyper-parameters as LAMB with grafting to understand the effect of
preconditioner. Moreover, step time can be optimized considerably as the current
implementation is not heavily optimized. For example, larger batch sizes help
amortize the preconditioning overhead, and reduce overall wall time to reach the
same accuracy. Furthermore, in our current implementation, all TPU cores compute
all the preconditioning statistics and the preconditioned gradients, which
involves over a hundred $1024\times 1024$ matrix multiplications. This repeated
work can be avoided by cross-replica sharding of weight update
\citep{xu2020automatic}, which distributes this computation across cores, and
should save at least half the step time overhead. Baseline results with LAMB optimizer used highly tuned learning rates. No tuning was carried out for Shampoo other than grafting the layer wise learning rates from LAMB.
\subsection{CIFAR-10}
\label{sec:cifar-details}
We train a ResNet-50 model on CIFAR-10 \citep{krizhevsky2009learning} with 2 cores of CloudTPU-v2 at batch size 2048. Our baseline achieves 93.45\% accuracy at 300 epochs, where as Shampoo reaches the same accuracy in 143 epochs. We  see an overall training time reduction of 42\% (1428 seconds to 827 seconds). As it is a smaller problem, the time taken for preconditioner inverse computation for the largest preconditioning matrix is less than 1ms on the CPU. We use a total of 8 CPU cores to run these inverses.

\subsection{Detailed results for experiments}
Approximate wall clock times for the various tasks are as follows:

\begin{tabular}{l|l|c|c}
\hline
Task&Model& Baseline & Shampoo\\
\hline
Recommendations: Criteo-1Tb& DLRM& 13 min& 8.2 min\\
Translation: WMT-14 En-Fr& Transformer& $\approx$ 12 hrs & 6.5 hrs\\
Translation: WMT-14 En-Fr& Transfomer-Big & $\approx$ 47 hrs & 29.5 hrs\\
Language Modeling: Wikipedia+Books& BERT-Large & 228 mins & 219 mins\\
\hline
\end{tabular}

\subsection{Breakdown of step-time in \cref{fig:en_fr_latency}}
\label{sec:steptime}
Each step of training consists of the following phases, whose times are shown in~\cref{fig:en_fr_latency}. 
\begin{itemize}
\item Forward Pass: Each core independently computes the predictions for each training example in its sub-batch.
\item Gradient: The gradient is for the sub-batch is computed using the back-propagation algorithm.
\item All reduction: The gradients for the sub-batches from all cores are averaged to compute the gradient for the minibatch. This is then sent back to each core.
\item Preconditioner statistics: The preconditioner statistics for adaptive algorithms are updated, e.g. for AdaGrad, we set $H_i := H_i + g_i^2$ for all parameters, while for Shampoo, we set $L_i := L_i + GG\tr$ etc.
\item Preconditioned gradient: The preconditioned gradient is computed - e.g. for AdaGrad, we compute $g_i / \sqrt{H_i}$, while for Shampoo, we compute $L^{-\sfrac{1}{4}}GR^{-\sfrac{1}{4}}$.
\item Parameter updates: The parameters are updated using the preconditioned gradients. This step is the same for all algorithms: $W := W - \eta \tilde{G}$, where $\tilde{G}$ 
is the preconditioned gradient. 
\end{itemize}
Note that the Shampoo computation of the preconditioners $L^{-\sfrac{1}{4}}, R^{-\sfrac{1}{4}}$ is pipelined on the host CPU, so does not show up in the step times.
\end{document}